\documentclass[12pt]{article}
\usepackage{latexsym,amsmath,amsthm,amssymb,amscd,xy,epsfig,graphicx,epstopdf}
\usepackage[round]{natbib}
\usepackage{apalike}
\usepackage{mathptmx}
\usepackage[mathscr]{euscript}
\usepackage{bigints}
\usepackage{caption}
\usepackage{float}
\usepackage{color}
\usepackage{fullpage}
\usepackage[title]{appendix}
\usepackage[pdftex]{hyperref}
\usepackage{algorithm}
\usepackage{algpseudocode}

\date{}\title{Weighted Clustering Ensemble: A Review}
\usepackage{authblk}
\author{\normalsize{Mimi Zhang}
}
\affil{\small{School of Computer Science and Statistics, Trinity College Dublin, Dublin 2, Ireland}}

\begin{document}
\maketitle

\begin{abstract}
Clustering ensemble, or consensus clustering, has emerged as a powerful tool for improving both the robustness and the stability of results from individual clustering methods. Weighted clustering ensemble arises naturally from clustering ensemble. One of the arguments for weighted clustering ensemble is that elements (clusterings or clusters) in a clustering ensemble are of different quality, or that objects or features are of varying significance. However, it is not possible to directly apply the weighting mechanisms from classification (supervised) domain to clustering (unsupervised) domain, also because clustering is inherently an ill-posed problem. This paper provides an overview of weighted clustering ensemble by discussing different types of weights, major approaches to determining weight values, and applications of weighted clustering ensemble to complex data. The unifying framework presented in this paper will help clustering practitioners select the most appropriate weighting mechanisms for their own problems.
\end{abstract}

\section{Introduction}
Clustering algorithms seek to partition data into clusters, or groups, according to certain similarity measures. The overall goal is to place similar data points in the same cluster, and dissimilar data points in different clusters. Clustering results can be either ``hard'' or ``fuzzy''. There exist a large number of clustering methods. These methods are characterized by different ways of measuring homogeneity, diverse procedures for searching the optimum partition, and various problem-dependent restrictions; see an overview by \cite{JAIN2010651}. However, it is well known that results from a single clustering method can vary due to a number of factors.
\begin{itemize}
\setlength{\itemsep}{-0.1cm}
  \item Variability due to local optimality: For NP-complete clustering algorithms, heuristic approaches have to be employed. Heuristic algorithms usually terminate after finding a locally optimal solution which can differ for multiple runs of the same algorithm.
  \item Variability due to algorithm: Objectives of different clustering algorithms are different. It is expected to have different clustering results for different algorithms.
  \item Variability due to data: In certain situations, different datasets may describe the same object, e.g., two different images of the same object under different illumination conditions and/or angles. It is possible that clustering results will be different even for the same object.
\end{itemize}
These inconsistencies motivate the development of clustering ensemble methods. Given an ensemble of base clusterings, the main objective of clustering ensemble is to extract a consensus clustering that maximizes certain objective function (the consensus function) defined by taking into account different information available from the given ensemble.
Building up an ensemble of base clusterings can be addressed by various ways, such as using different subsets of features, using different subsets of objects, varying one or more parameters of the clustering algorithm, or using different clustering algorithms. The consensus function is the primary step in any clustering ensemble algorithm. Precisely, the great challenge in clustering ensemble is the definition of an appropriate consensus function, capable of improving the results of single clustering algorithms. \cite{VEGA-PONS} divide existing consensus functions into two categories. In the first category, the consensus clustering is obtained by analyzing objects co-occurrence: how many times an object belongs to one cluster or how many times two objects belong to the same cluster. For example, the evidence accumulation based method \citep{Fred2005CombiningMC} falls in this category. In the second category, the consensus clustering is called the median partition which has the maximal similarity with all partitions in the ensemble. For example, the nonnegative matrix factorization (NMF) based method \citep{DingNMF} falls in this category. \cite{ghaemi2009survey} provides another taxonomy of consensus functions. \cite{Xanthopoulos2014} gives a short survey on exact and approximating clustering ensemble algorithms. \cite{BOONGOEN20181} divide consensus functions into four categories: direct, feature-based, pairwise-similarity based and graph-based approaches.

Most of the clustering ensemble algorithms give every base clustering the same weight. However, not all clusterings in the ensemble have the same quality. Different base clusterings could differ significantly or could be highly correlated. When base clusterings differ significantly, the consensus by simply averaging is a brute-force voting. The brute-force voting by highly divergent parties is not stable. On the other hand, when individual clusterings are highly correlated, these redundant clusterings will tend to bias the consensus clustering towards these correlated clusterings. Therefore, a simple average of all clusterings may not be the best choice, which motivates the development of weighted clustering ensemble. An intuitive idea of weighted clustering ensemble is to give a weight to each base clustering according to its quality/diversity in the clustering ensemble. \cite{BERIKOV2017427} theoretically investigate a weighted clustering ensemble method in which the consensus clustering is obtained by applying the average linkage agglomerative clustering to the weighted similarity matrix $\mathbf{B}\mathbf{W}\mathbf{B}'$ (see Section \ref{BP}). The clustering ensemble is created by applying a single algorithm using parameters taken at random, and the weights are determined by an arbitrary evaluation function. It is proved that under certain natural assumptions, the misclassification probability for any pair of objects converges to zero as the ensemble size increases. They point out that weighting the base clusterings in constructing the consensus clustering is particularly essential when the ensemble size should be reasonably small due to the time and storage restrictions.

Other than assigning clustering weights, efforts have been made to design weights for clusters, features, and data points/objects. Many stability measures have been proposed to validate clusterings. It is likely that a clustering containing one or more high-quality clusters is adjudged unstable by a stability measure, and as a result, is completely neglected. Hence, instead of treating equally all the clusters of a clustering, one can design cluster-level weights. The primary motivation for weighting clusters is to give lower weights to clusters that are unstable. To dodge the curse of dimensionality, many different subspace clustering methods have been proposed. In high dimensional spaces, it is highly likely that, for any given pair of objects within the same cluster, there exist at least a few dimensions on which the objects are far apart from each other. As a consequence, distance functions that equally use all input features may not be effective. Different weighted clustering ensemble algorithms have been developed to determine the optimal weight for each feature. The primary motivation for weighting objects is to give higher weights to objects that are hard to cluster. The success of boosting algorithms for classification tasks motivates the believe that boosting a clustering algorithm can lead to a more accurate consensus clustering. This is accomplished by adapting the (sampling) weights of objects according to previous base clusterings. In addition to the difference in types of weights, the approach to determine weight values differs as well. There are mainly two approaches to determine weight values: weights are either calculated from the given clustering ensemble using specific validation criteria, or treated as variables and determined by solving an optimization problem. We call these two approaches the fixed-weight approach and the variable-weight approach. The main purpose of this paper is to compile and analyze state-of-the-art methods for weighted clustering ensemble. This work also includes different applications of weighted clustering ensemble, with several research issues and challenges being highlighted.

The rest of this paper is organized as follows. Section \ref{BP} introduces two main consensus function approaches and two main weight determination approaches. Sections \ref{fixedweigths} and \ref{variableweights} review the weighted clustering ensemble research on fixed weights and variable weights, respectively. Each section is divided into multiple subsections, with one subsection addressing one weight type. Finally, Appendix \ref{application} explores the applications of weighted clustering ensemble methods to multi-view data and temporal data.

\section{Preliminaries}\label{BP}
Denote $\mathbb{X}=\{\pmb{x}_1, \ldots, \pmb{x}_n\}$ the set of objects/data points, where each $\pmb{x}_i$ ($i=1, \ldots, n$) is described by $p$ features. Let $\mathscr{C}=\{C_1, \ldots, C_M\}$ denote the ensemble of $M$ base clusterings, with the $m$th ($m=1, \ldots, M$) clustering containing $k_m$ clusters: $C_m=\{\mathbb{C}_m^1, \ldots, \mathbb{C}_m^{k_m}\}$. In the current work, unless otherwise stated, all clusterings are hard (crisp) clusterings. For any $m$ and $v=1, \ldots, k_m$, let $\pmb{c}_m^v$ denote the centroid of cluster $\mathbb{C}_m^v$. Denote $\mathscr{C}_{\mathbb{X}}$ the set of all possible clusterings on the data $\mathbb{X}$ ($\mathscr{C}\subset\mathscr{C}_{\mathbb{X}}$). The objective of clustering ensemble is to combine the $M$ base clusterings into a single consolidated clustering $C^*$, called the consensus clustering.

For example, in the median partition approach, $C^*$ is obtained from $C^*=\arg\max\limits_{C\in\mathscr{C}_{\mathbb{X}}}\sum_{m=1}^{M}\phi(C, C_m)$, where the function $\phi(\cdot,\cdot)$ measures the similarity between two clusterings. In the previous equation, all base clusterings are treated equally. As a natural generalization, we can assign a weight to each base clustering and obtain the consensus clustering by solving:
\begin{equation}
C^*=\arg\max\limits_{C\in\mathscr{C}_{\mathbb{X}}}\sum_{m=1}^{M}w_m\phi(C, C_m),
\label{weighted}
\end{equation}
where $w_m\geq0$ and $\sum_{m=1}^{M}w_m=1$.

The similarity function $\phi(\cdot,\cdot)$ has the range $[0, 1]$, with a unity value implying the maximum agreement between two clusterings and a zero value implying no agreement. For readability, we here exemplify the similarity function via the normalized mutual information (NMI) measure. Given two clusterings $C=\{\mathbb{C}^1, \ldots, \mathbb{C}^{k}\}\in\mathscr{C}_{\mathbb{X}}$ and $C_m=\{\mathbb{C}_m^1, \ldots, \mathbb{C}_m^{k_m}\}$, let $n^v=|\mathbb{C}^v|$ be the number of objects in cluster $\mathbb{C}^v (\in C)$, $n_m^r=|\mathbb{C}_m^r|$ the number of objects in cluster $\mathbb{C}_m^r (\in C_m)$, and $n_{vr}=|\mathbb{C}^v\cap \mathbb{C}_m^r|$ the number of objects in both $\mathbb{C}^v$ and $\mathbb{C}_m^r$ ($v=1, \ldots, k$ and $r=1, \ldots, k_m$). Table \ref{CT}
\begin{table}[!ht]
  \centering
  \caption{The contingency table for clusterings $C=\{\mathbb{C}^1, \ldots, \mathbb{C}^{k}\}$ and $C_m=\{\mathbb{C}_m^1, \ldots, \mathbb{C}_m^{k_m}\}$.}\label{CT}
  \begin{tabular}{llllll}
    \hline
            &$\mathbb{C}_m^1$   &$\mathbb{C}_m^2$   &\ldots &$\mathbb{C}_m^{k_m}$ &sum\\
    $\mathbb{C}^1$ &$n_{11}$  &$n_{12}$  &\ldots &$n_{1k_m}$  &$n^1$ \\
    $\mathbb{C}^2$ &$n_{21}$  &$n_{22}$  &\ldots &$n_{2k_m}$  &$n^2$ \\
    \ldots  &\ldots    &\ldots    &\ldots &\ldots      &\ldots \\
$\mathbb{C}^{k}$ &$n_{k1}$&$n_{k2}$&\ldots &$n_{kk_m}$&$n^{k}$ \\
        sum &$n_m^1$   &$n_m^2$   &\ldots &$n_m^{k_m}$ &$n$ \\
    \hline
  \end{tabular}
\end{table}
is a contingency table showing the overlap between different clusters. The entropy associated with clustering $C$ is $H(C)=-\sum_{v=1}^k\frac{n^v}{n}\log(\frac{n^v}{n})$, and that with clustering $C_m$ is $H(C_m)=-\sum_{r=1}^{k_m}\frac{n_m^r}{n}\log(\frac{n_m^r}{n})$. The joint entropy of $C$ and $C_m$ is $H(C, C_m)=-\sum_{v}\sum_{r}\frac{n_{vr}}{n}\log(\frac{n_{vr}}{n})$. The mutual information for measuring the similarity between clusterings $C$ and $C_m$ is $I(C, C_m)=H(C)+H(C_m)-H(C, C_m)$. While different versions of NMI have been reported \citep{Kvalseth1987,Yao2003, Strehl:2003}, in the following, we let $\phi(\cdot, \cdot)$ denote the NMI defined by \cite{Strehl:2003} due to its ubiquitousness. In the current work, function $H(\cdot)$ is reserved for calculating entropy; its argument can be either a clustering or a set of probabilities summed to 1. For example, $H(C)=H(\frac{n^1}{n}, \ldots, \frac{n^k}{n})=-\sum_{v=1}^k\frac{n^v}{n}\log(\frac{n^v}{n})$.

\begin{figure}[!ht]
  \centering
  \includegraphics{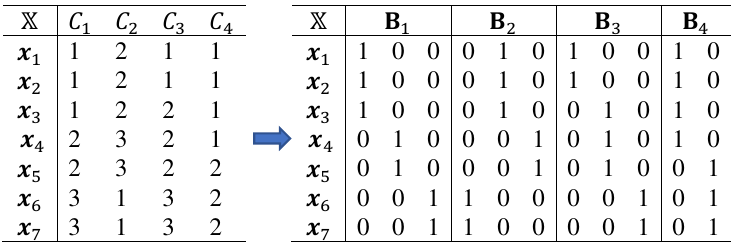}
  \caption{Representing base clusterings by binary membership indicator matrices. The numbers \{1, 2, 3\} in each column of the left table are cluster indicators. Each column in the binary membership indicator matrix $\mathbf{B}_m$  corresponds to a cluster in the base clustering $C_m$, $m=1, \ldots, 4$.}
\label{BinaryMatrix}
\end{figure}

The clustering-weighting idea can be readily extended to clustering ensemble methods based on objects co-occurrence. We illustrate the idea via the cluster-based similarity partitioning algorithm (CSPA) proposed by \cite{Strehl:2003}. Particularly, for each base clustering $C_m$, we can construct a binary membership indicator matrix $\mathbf{B}_m$; see Figure \ref{BinaryMatrix}.
Let $\mathbf{B}^*$ denote the binary membership indicator matrix for the consensus clustering $C^*$. Each column in a binary membership indicator matrix represents a cluster. Then the co-occurrence matrix for clustering $C_m$ is simply $\mathbf{B}_m\mathbf{B}_m'$, where the prime represents the transpose operator. The relation $[\mathbf{B}_m\mathbf{B}_m']_{ij}=1$ ($1\leq i,j\leq n$) indicates that objects $\pmb{x}_i$ and $\pmb{x}_j$ are in the same cluster of clustering $C_m$. The concatenated matrix $\mathbf{B}=[\mathbf{B}_1, \mathbf{B}_2, \ldots, \mathbf{B}_M]$ is the binary membership indicator matrix for the ensemble $\mathscr{C}$. Then the similarity matrix (or, co-association matrix) for the ensemble $\mathscr{C}$ is $\frac{1}{M}\mathbf{B}\mathbf{B}'$. The element $[\frac{1}{M}\mathbf{B}\mathbf{B}']_{ij}$ denotes the fraction of clusterings in which objects $\pmb{x}_i$ and $\pmb{x}_j$ belong to the same cluster. We let $G(\frac{1}{M}\mathbf{B}\mathbf{B}')$ denote the graph constructed from $\frac{1}{M}\mathbf{B}\mathbf{B}'$: each vertex represents an object, edge weights are the elements in the similarity matrix $\frac{1}{M}\mathbf{B}\mathbf{B}'$, and no edge exists between objects $\pmb{x}_i$ and $\pmb{x}_j$ if $[\frac{1}{M}\mathbf{B}\mathbf{B}']_{ij}=0$. The graph $G(\mathbf{B}_m\mathbf{B}_m')$ for clustering $C_m$ can be defined alike. The CSPA applies the graph partitioning algorithm METIS on the graph $G(\frac{1}{M}\mathbf{B}\mathbf{B}')$ to obtain the consensus clustering. To account for clustering quality, we might assign a weight $w_m\geq0$ to clustering $C_m$ and define a diagonal matrix of weights: $\mathbf{W}=\mbox{diag}(\underbrace{w_1, \ldots, w_1}\limits_{k_1}, \ldots, \underbrace{w_M, \ldots, w_M}\limits_{k_M})$ and $\sum_{m=1}^{M}w_m=1$. Then a more stable consensus clustering can be obtained by applying a graph partitioning algorithm on $G(\mathbf{B}\mathbf{W}\mathbf{B}')$.

There are mainly two approaches to determine the clustering weights $\{w_1, \ldots, w_M\}$. In the first approach, weights are related to various clustering validation indices. A plethora of clustering validation criteria have been developed for characterizing different properties of a clustering. For example, for a clustering $C=\{\mathbb{C}^1, \ldots, \mathbb{C}^{k}\}$ with centroids $\{\pmb{c}^1, \ldots, \pmb{c}^k\}$, the Dunn's validity index (DVI) is defined as: $\min_{1\leq r\neq v\leq k}\{\frac{\mbox{set distance between }\mathbb{C}^r \mbox{ and } \mathbb{C}^v}{\max\limits_{r=1, \ldots, k}\{\mbox{diameter of }\mathbb{C}^r\}}\}$, while the modified Hubert statistic is defined as:
\begin{equation*}
\frac{2}{n(n-1)}\sum_{i=1}^{n-1}\sum_{j=i+1}^{n}\|\pmb{x}_i-\pmb{x}_j\|_2 \sum_{r=1}^{k}\sum_{v=1}^{k}\|\pmb{c}^r-\pmb{c}^v\|_2\delta(\pmb{x}_i\in\mathbb{C}^r)\delta(\pmb{x}_j\in\mathbb{C}^v).
\end{equation*}
$\delta(\cdot)$ is the indicator function which equals 1 if the argument is true and 0 otherwise. Here and in the following, we let $\|\cdot\|_q$ denote the $L_q$ norm ($q=1$ and 2), and $\|\cdot\|_F$ the Frobenius norm.
In the second approach, both the weights $\{w_1, \ldots, w_M\}$ and the consensus clustering are treated as optimization variables. However, a potential problem lurks in that optimization problems may be ill-posed: optimal solutions will put the unity weight on a single base clustering. In such cases, one can impose certain regularization on the weights $\{w_1, \ldots, w_M\}$. In the current work, we always use $\lambda (\geq0)$ (sometimes with subscript) to dictate the amount of regularization/penalization. Global optimization is often intractable, and typically heuristic algorithms are developed in which the weights $\{w_1, \ldots, w_M\}$ and the consensus clustering are optimized iteratively.

\begin{figure}[!ht]
  \centering
  \includegraphics[scale=0.7]{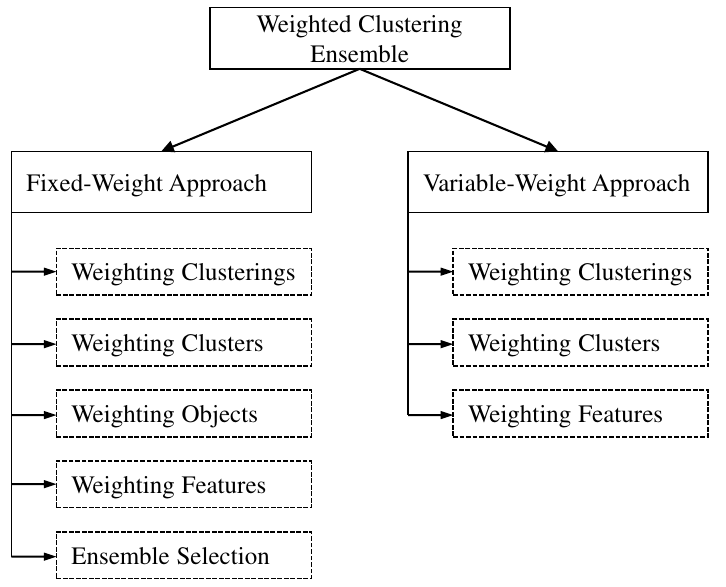}
  \caption{Taxonomy of methods for weighted clustering ensemble.}
\label{taxonomy}
\end{figure}

Apart from clustering weights, efforts have been made to design weights for features, clusters, and objects. This review also includes research studies on ensemble selection, with the interpretation that the un-selected elements (e.g., clusterings) are assigned the weight zero. Figure \ref{taxonomy}
provides an overview of weighted clustering ensemble taxonomy following the two weight-determination approaches described before. Table \ref{notations} summarizes notations we will use throughout the paper.
\begin{table}[!ht]
  \centering
  \caption{Notations adopted throughout the paper.}\label{notations}
  \begin{tabular}{ll}
    \hline
    $\mathbb{X}=\{\pmb{x}_1, \ldots, \pmb{x}_n\}$& the set of objects/data points\\
    $\mathscr{C}=\{C_1, \ldots, C_M\}$& the ensemble of $M$ base clusterings\\
    $C_m=\{\mathbb{C}_m^1, \ldots, \mathbb{C}_m^{k_m}\}$& the $m$th base clustering with $k_m$ clusters ($m=1, \ldots, M$)\\
    $\pmb{c}_m^v$& the centroid of cluster $\mathbb{C}_m^v$ ($\mathbb{C}_m^v\in C_m$, $v=1, \ldots, k_m$)\\
    $C=\{\mathbb{C}^1, \ldots, \mathbb{C}^{k}\}$& a clustering with $k$ clusters\\
    $\mathscr{C}_{\mathbb{X}}$& the set of all possible clusterings on the data $\mathbb{X}$ ($\mathscr{C}\subset\mathscr{C}_{\mathbb{X}}$)\\
    $C^*$& the consensus clustering\\
    $\{w_1, \ldots, w_M\}$& clustering weights\\
    $\phi(\cdot,\cdot)$& the normalized mutual information (NMI) function\\
    $H(\cdot)$& the entropy function\\
    $\mathbf{B}_m$& the binary membership indicator matrix for the base clustering $C_m$\\
    $\mathbf{B}^*$& the binary membership indicator matrix for the consensus clustering $C^*$\\
    $\mathbf{B}_m\mathbf{B}_m'$& the co-occurrence matrix for clustering $C_m$\\
    $\mathbf{B}=[\mathbf{B}_1, \mathbf{B}_2, \ldots, \mathbf{B}_M]$& the binary membership indicator matrix for the ensemble $\mathscr{C}$\\
    $\frac{1}{M}\mathbf{B}\mathbf{B}'$& the similarity/co-association matrix for the ensemble $\mathscr{C}$\\
    $G(\cdot)$& the graph constructed from a co-occurrence or co-association matrix\\
    $\mathbf{W}$& the diagonal matrix of weights $\mathbf{W}=\mbox{diag}(\underbrace{w_1, \ldots, w_1}\limits_{k_1}, \ldots, \underbrace{w_M, \ldots, w_M}\limits_{k_M})$\\
    $\delta(\cdot)$& the 0-1 indicator function\\
    $\|\cdot\|_q$~\& $\|\cdot\|_F$& the $L_q$ norm and Frobenius norm\\
    $\lambda (\geq0)$& the regularization/penalization parameter\\
    $i\cdot$ (resp. $\cdot j$)& the $i$th row (resp. $j$th column) of a matrix\\
    \hline
  \end{tabular}
\end{table}

\section{Fixed-Weight Approach}\label{fixedweigths}
\subsection{Weighting Clusterings}
In \cite{ZHOU200677}, the weight for a clustering $C\in\mathscr{C}$ is the normalized value of $(M-1)/\sum_{C_m\neq C}\phi(C, C_m)$, namely, the inverse of the averaged NMI. The motivation is that the larger the averaged NMI is, the less information revealed by $C$ has not been contained by the other clusterings. If a weight $w_m$ is smaller than the predetermined threshold $\frac{1}{M}$, clustering $C_m$ will be excluded from the ensemble. Then the consensus clustering is obtained via weighted voting by the remaining clusterings. Before voting, clustering labels need be aligned. The labeling correspondence problem is solved by (1) randomly selecting a clustering from the ensemble as the reference clustering; (2) matching a cluster from a clustering $C$ to the cluster in the reference clustering that shares the most objects; (3) repeating step (2) until all the clusters in $C$ are matched. \cite{Zhangetal2009} apply the $k$-means algorithm on different subsets of $\mathbb{X}$ to generate the clustering ensemble. To solve the labeling correspondence problem, the reference clustering is randomly selected from the ensemble. The labels of each base clustering are aligned w.r.t. the reference clustering by minimizing the sum of the distances between paired centroids (one from the base clustering and the other from the reference clustering). After relabelling, the consensus clustering is obtained by weighted voting where the weight for a clustering is the NMI (sqrt) between itself and the reference clustering. For clustering massive data, \cite{6888693} propose a hybrid sampling scheme: in each iteration, a subset of objects is sampled from $\mathbb{X}$ without replacement, and then a bootstrap sample of size $n$ is generated from the subset with replacement. The $k$-means algorithm is applied on the bootstrap samples to generate the base clusterings, and, using the weights from \cite{ZHOU200677}, the consensus clustering is obtained by weighted voting. To classify hyperspectral images, \cite{Alhichri2014} run the fuzzy c-means algorithm several times with different initializations and different subsets of features to create the clustering ensemble. The labeling correspondence problem is solved by the Hungarian method, where the reference clustering is the one having the largest entropy. The consensus clustering is obtained by applying the Markov random field method on the re-labeled base clusterings, where each clustering is weighted by the averaged NMI (sqrt) value.

\cite{Duarte2006WeightedEA} employ 16 clustering validation criteria in order to determine the weight of each base clustering. The validation indices are: raw and normalized Hubert statistics, Dunn's validation index, Davies-Bouldin validation index, Calinski-Harabasz validation index, Silhouette validation index, squared error index, R-squared index and root-mean-square standard error, SD and S\_Dbw validation indices, index I, Xie-Beni validation index, Krzanowski-Lai validation index, Hartigan validation index, and point symmetry index. The $M$ base clusterings are obtained by applying a clustering algorithm on $M$ randomly generated subsets of $80\%$ objects from $\mathbb{X}$. The weight $w_m$ for clustering $C_m$ is $w_m=\frac{1}{16}\sum_{v=1}^{16}\mbox{``normalized value of the $v$th validation index for clustering $C_m$''}$. The weighted similarity matrix $\mathbf{B}\mathbf{W}\mathbf{B}'$ is adjusted to account for the number of times that two objects occur in the same subset:
\begin{equation*}
[\mathbf{B}\mathbf{W}\mathbf{B}']_{ij}\leftarrow\frac{[\mathbf{B}\mathbf{W}\mathbf{B}']_{ij}}{\mbox{number of times that objects $\pmb{x}_i$ and $\pmb{x}_j$ occur in the same subset}}.
\end{equation*}
The consensus clustering is obtained by applying a clustering algorithm, e.g., the $k$-means algorithm, on the adjusted weighted similarity matrix. The clustering weights in \cite{Unlu2017} are determined by an individual internal clustering validation criterion. The authors suggest to use the Silhouette validation index, Calinski-Harabasz validation index or Davies-Bouldin validation index. To obtain the consensus clustering, they apply the CSPA on the graph $G(\mathbf{B}\mathbf{W}\mathbf{B}')$.
In real practice, we often do not know which clustering validation indices to use for our data. \cite{Ilc2020} proposes to apply feature selection/extraction methods to remove redundant clustering validation indices from a committee. Firstly,  each base clustering in the ensemble is validated by a committee of $J$ clustering validation indices, giving an $M\times J$ matrix $\mathbf{V}$, where $\mathbf{V}_{mj}$ is the value of the $j$th validation index for the $m$th base clustering. The values in the matrix $\mathbf{V}$ are then unified to have the same scale, and a larger value indicates a better clustering. To filter redundant validation indices, the author investigates different feature selection/extraction methods (e.g., spectral feature selection and probabilistic principal component analysis). By applying a feature selection/extraction method on $\mathbf{V}$, we obtain a filtered matrix $\ddot{\mathbf{V}}$ that has fewer columns than $\mathbf{V}$. Finally, the clustering weight $w_m$ is calculated by aggregating the row values $\ddot{\mathbf{V}}_{m\cdot}$, and the consensus clustering is obtained by applying a clustering algorithm (e.g., CSPA) on the weighted similarity matrix $\frac{1}{M}\mathbf{B}\mathbf{W}\mathbf{B}'$.

To make multivariate regression trees applicable in the traditional clustering framework where there are no response variables, one can replicate the $p$ features as the response variables. When $p$ is too large, the response space dimension can be reduced via either principal component analysis or factor analysis. In \cite{Smyth2007}, $M$ multivariate regression trees are built from $M$ different subsets of predictors, where each subset of predictors is a random sampling of the $p$ features with percentage 5\%. The terminal nodes of the $m$th tree are the clusters of clustering $C_m$. Tree weights (i.e., clustering weights) are determined by the forward stagewise linear regression (with the objective of obtaining a boosted tree model). The motivation of employing the forward stagewise linear regression is to enforce parsimony on the clustering weights $\{w_1, \ldots, w_M\}$. Finally, they propose a similarity-based $k$-means algorithm to obtain the consensus clustering from the weighted similarity matrix $\mathbf{B}\mathbf{W}\mathbf{B}'$.

In \cite{Vega-Pons2008}, clustering weights are determined by a set of property indices $\{I_1, \ldots, I_{\mathbf{i}}\}$, where $I_v(C_m)$ measures the degree of a property accomplished by clustering $C_m$ ($v=1, \ldots, \mathbf{i}$). The four property indices adopted by the authors are: inter-cluster distance, intra-cluster distance, mean size of clusters, and difference between cluster sizes. The weight $w_m$ for clustering $C_m$ is
\begin{equation*}
w_m=\sum_{v=1}^{\mathbf{i}}H(\frac{I_v(C_1)}{\sum_{j=1}^{M}I_v(C_j)}, \ldots, \frac{I_v(C_M)}{\sum_{j=1}^{M}I_v(C_j)})(1-|I_v(C_m)-\frac{1}{M}\sum_{j=1}^{M}I_v(C_j)|).
\end{equation*}
The higher value of $w_m$, the stronger likeness among the $\{I_v(C_1), \ldots, I_v(C_M)\}$ values. The objective is to solve problem (\ref{weighted}) in which the similarity function $\phi(\cdot, \cdot)$ is explained as follows. Let $h^*$ denote a path of distinct vertices in graph $G(\mathbf{B}^*\mathbf{B}^{*'})$, and $h_m$ a path of distinct vertices in graph $G(\mathbf{B}_m\mathbf{B}_m')$. The similarity function is defined as
\begin{equation*}
\phi(C^*, C_m)=\sum_{h^* \in G(\mathbf{B}^*\mathbf{B}^{*'})}\sum_{h_m \in G(\mathbf{B}_m\mathbf{B}_m')}\delta(h^*=h_m)\Pr(h^*)\Pr(h_m),
\end{equation*}
where, e.g., $\Pr(h^*)$ is the probability of a random walk along the path $h^*$. They prove that $\phi(\cdot, \cdot)$ is a positive semi-definite kernel, and hence problem (\ref{weighted}) can be transformed into a simple optimization problem in the corresponding reproducing kernel Hilbert space, where the optimal solution can be readily obtained. However, the mapping of the optimal solution in the reproducing kernel Hilbert space back into $\mathscr{C}_{\mathbb{X}}$ is computationally prohibitive. The authors approximate $C^*$ by solving another optimization problem.
In cases where the user knows what characteristics are considered as good, the user may want to maximize those characteristics. Hence, \cite{VEGAPONS20102712} define another type of clustering weights: $w_m=\sum_{v=1}^{\mathbf{i}}(1-|I_v(C_m)-\max_{j=1, \ldots, M} I_v(C_j)|)$.  The weight assigned to each clustering measures how close the property indices of a clustering are to their maximum values. The four property indices adopted by the authors are: Dunn's validation index, Silhouette validation index, compactness of the clusters, and connectedness of the clusters. The similarity function is defined as
\begin{equation*}
\phi(C, C_m)=\frac{\sum_{S\subseteq\mathbb{X}}\sum_{v=1}^k\sum_{r=1}^{k_m}\delta(S\subseteq \mathbb{C}^v)\delta(S\subseteq \mathbb{C}_m^r)\mu(S|C)\mu(S|C_m)}{\sqrt{\sum_{S\subseteq\mathbb{X}}\sum_{v=1}^k\delta(S\subseteq \mathbb{C}^v)\mu^2(S|C)}\sqrt{\sum_{S\subseteq\mathbb{X}}\sum_{r=1}^{k_m}\delta(S\subseteq \mathbb{C}_m^r)\mu^2(S|C_m)}},
\end{equation*}
where, e.g., $\mu(S|C)$ measures the significance of the subset $S$ for clustering $C$. The sum is over all possible subsets of $\mathbb{X}$, and the similarity function $\phi(\cdot, \cdot)$ is proved to be a positive semi-definite kernel. Then again the authors transform problem (\ref{weighted}) into an optimization problem in the corresponding feature space and obtain the optimal solution in the feature space. Finally the simulated annealing meta-heuristic is employed to solve the pre-image problem. \cite{VEGAPONS20112163} point out that, for clustering ensemble methods based on objects co-occurrence, the similarity matrix should summarize as much information as possible from the clustering ensemble, and that two objects belonging to the same cluster does not contribute with the same information for every clustering in the ensemble. Hence they define a matrix $\mathbf{R}$:
\begin{equation*}
\mathbf{R}_{ij}=\sum_{m=1}^{M}\sum_{r=1}^{k_m}\delta(\pmb{x}_i\in\mathbb{C}_m^r)\delta(\pmb{x}_j\in\mathbb{C}_m^r)\times\frac{|C_m|}{|\mathbb{C}_m^r|}\times\mbox{``similarity between $\pmb{x}_i$ and $\pmb{x}_j$ used to obtain $C_m$''}.
\end{equation*}
Here, $|C_m|$ is the number of clusters in $C_m$, while $|\mathbb{C}_m^r|$ is the number of objects in $\mathbb{C}_m^r$. Two approaches to obtain the consensus clustering from $\mathbf{R}$ are proposed. In the first approach, a hierarchical agglomerative clustering method is applied on $\mathbf{R}$ to obtain the consensus clustering. The second approach is similar to that in \cite{VEGAPONS20102712}, and the difference is that, when calculating the clustering weights, property indices are applied on the representation data $\{\mathbf{R}_{i\cdot}: i=1, \ldots, n\}$, instead of the original data $\mathbb{X}$. In the current work, the subscript ``$i\cdot$'' (resp. ``$\cdot j$'') associated with a matrix represents the $i$th row (resp. $j$th column) of the matrix. The representation data $\{\mathbf{R}_{i\cdot}: i=1, \ldots, n\}$ are real-valued, allowing the use of mathematical tools that may not be available for the original data.

The study in \cite{Gullo2009} proposes three weighting schemes. We here only explain the single weighting scheme. $\forall$ $C_m\in\mathscr{C}$, the diversity of the ensemble $\mathscr{C}\setminus\{C_m\}$ is defined as
\begin{equation*}
 d_m=\frac{2}{(M-1)(M-2)}\sum_{\substack{1\leq j<k\leq M\\j, k\neq m}}[1-\phi(C_j, C_k)].
\end{equation*}
The NMI measure in the above equation can also be replaced by the F-measure. Then the weight for clustering $C_m$ is $w_m=\alpha\frac{d_m}{\sum_{j=1}^{M}d_j}+(1-\alpha)\frac{N_{\mu, \sigma}(d_m)}{\sum_{j=1}^{M}N_{\mu, \sigma}(d_j)}$, where $\alpha$ is a user-defined parameter, and $N_{\mu, \sigma}(\cdot)$ is the Normal density function with mean $\mu$ and standard deviation $\sigma$. The Normal density is introduced to give higher weights to clusterings with moderate levels of diversity. To include the weights $\{w_1, \ldots, w_M\}$ into an instance-based clustering ensemble, the authors first replace each object $\pmb{x}_i\in\mathbb{R}^p$ with $\tilde{\pmb{x}}_i\in\mathbb{R}^M$, where $\tilde{x}_{im}$ depends on clustering $C_m$, then calculate a new pairwise distance matrix $\mathbf{R}$ (e.g., $\mathbf{R}_{ij}=\sqrt{\langle\pmb{w}, (\tilde{\pmb{x}}_i-\tilde{\pmb{x}}_j)^2\rangle}$), and finally perform a clustering task on the objects $\mathbb{X}$ using the new pairwise distance matrix $\mathbf{R}$. To include the weights $\{w_1, \ldots, w_M\}$ into a cluster-based clustering ensemble, the authors first perform a clustering task on all the clusters in the ensemble (using, e.g., the Jaccard coefficient as the pairwise distance), and then assign each object to a meta-cluster (using, e.g., weighted majority voting where the weight for a cluster is the weight for the clustering that contains the cluster). To include the weights $\{w_1, \ldots, w_M\}$ into a hybrid clustering ensemble, the authors first build a bipartite graph of which the edge set is comprised of links between objects and clusters, then multiply each original edge weight by $(1+w_m)$ where $w_m$ is the weight for the clustering that contains the cluster, and finally apply a graph partitioning algorithm on the re-weighted bipartite graph to obtain the consensus clustering.

\cite{6132997} define the score of a base clustering $C=\{\mathbb{C}^1, \ldots, \mathbb{C}^{k}\}$ as
\begin{equation*}
\frac{1}{k}\sum_{v=1}^{k} \frac{1}{|\mathbb{C}^v|}\sum_{\pmb{x}_i\in \mathbb{C}^v}\|\pmb{x}_i-\pmb{c}^v\|_2^2-\frac{2}{k(k-1)}\sum_{v=1}^{k}\sum_{r=1}^{k}\|\pmb{c}^v-\pmb{c}^r\|_2^2.
\end{equation*}
The first term quantifies the average intra-cluster similarity; the second term quantifies the average inter-cluster similarity. The scores are normalized over the base clusterings to obtain the clustering weights. A 0-1 binary matrix $\mathbf{I}$ is obtained by thresholding the weighted similarity matrix $\mathbf{B}\mathbf{W}\mathbf{B}'$: $\mathbf{I}_{ij}=\delta([\mathbf{B}\mathbf{W}\mathbf{B}']_{ij}>\mbox{a predetermined threshold})$. The consensus clustering is obtained by grouping the rows and columns of $\mathbf{I}$, such that the matrix-blocks defined by the row-groups and column-groups are as homogeneous as possible. Specifically, suppose that the rows of $\mathbf{I}$ are grouped into $k$ different clusters. Since $\mathbf{I}$ is symmetric, the columns of $\mathbf{I}$ are grouped in the same way. Then the rows and columns of $\mathbf{I}$ are permuted according to the grouping, resulting in $k\times k$ matrix-blocks in the permuted $\mathbf{I}$. The objective is to determine an optimal grouping (i.e., the consensus clustering) that maximizes the total homogeneity: $$\sum_{j=1}^{k\times k}-H(\frac{\mbox{number of 0's in the $j$th block}}{\mbox{size of the $j$th block}}, ~\frac{\mbox{number of 1's in the $j$th block}}{\mbox{size of the $j$th block}}).$$

For semi-supervised clustering, \cite{YANG2012101} generalize the $k$-means algorithm to incorporate prior knowledge. Prior knowledge introduces constraints in the form of whether a pair of objects must be in the same group (must-link) or in different groups (cannot-link). The generalized algorithm assigns an object (say, $\pmb{x}_i$) to the cluster in which $\pmb{x}_i$ has a must-link object, or to the closest cluster in which $\pmb{x}_i$ has no cannot-link objects. If no such cluster is found, then $\pmb{x}_i$ is assigned to its closest cluster (say, $\mathbb{C}^v$). If $\pmb{x}_i$ has a cannot-link object in $\mathbb{C}^v$ (say, $\pmb{x}_j$), then $\pmb{x}_j$ is moved out of $\mathbb{C}^v$; $\pmb{x}_j$ will be reassigned to a cluster other than $\mathbb{C}^v$ in the same way as assigning $\pmb{x}_i$. By multiple runs of the generalized algorithm, a clustering ensemble $\mathscr{C}=\{C_1, \ldots, C_M\}$ is obtained. The weight for clustering $C_m$ is the normalized value of $\frac{1}{M}\sum_{C\in\mathscr{C}}\phi(C, C_m)$, and the weighted similarity matrix $\mathbf{B}\mathbf{W}\mathbf{B}'$ is obtained. Apparently, the performance of the generalized algorithm is sensitive to the assignment order of the objects. Hence, utilizing the weighted similarity matrix, the authors define the ``certainty'' of an object: $\mbox{certainty}(\pmb{x}_i)=\frac{1}{n-1}\sum_{j=1, j\neq i}^{n}|1-[2\mathbf{B}\mathbf{W}\mathbf{B}']_{ij}|$. A new clustering ensemble is created by multiple runs of the generalized algorithm where objects are ordered according to their certainty values. To obtain a consensus clustering satisfying the must-link and cannot-link constraints, the authors modify the self-organizing map in the same manner as they modify the $k$-means algorithm. The modified self-organizing map is applied to the new clustering ensemble to obtain the consensus clustering. Given a priority ranking of the constraints, \cite{okabe2013uncertainty} employ Cop-Kmeans \citep{Wagstaff:2001:CKC} to produce a base clustering, represented by an indicator matrix $\pmb{K}^m$: $\pmb{K}_{ij}^m=\delta(\pmb{x}_i \mbox{ and } \pmb{x}_j \mbox{ are in the same cluster})-\delta(\pmb{x}_i \mbox{ and } \pmb{x}_j \mbox{ are in different clusters})$, where $m$ is the iteration number. Then in the $(m+1)$st iteration, the priority of the $l$th constraint, denoted by $\alpha_l^{m+1}$, is $\alpha_l^{m+1}=\alpha_l^m\exp(-w_m[\delta(\pmb{x}_i \mbox{ and } \pmb{x}_j \mbox{ must link})-\delta(\pmb{x}_i \mbox{ and } \pmb{x}_j \mbox{ cannot link})]\pmb{K}_{ij}^m)$, assuming that the $l$th constraint is related to objects $\pmb{x}_i \mbox{ and } \pmb{x}_j$. Here, $w_m$ is the proportion of unsatisfied constraints in the $m$th iteration. The priority values determine the ranking of the constraints in the $(m+1)$st iteration. The final consensus clustering is obtained by applying the kernel $k$-means algorithm to the weighted indicator matrix $\sum_{m=1}^{M}w_m\pmb{K}^m$.

\cite{BERIKOV201499} assumes that there are $M$ different algorithms for partitioning the data $\mathbb{X}$, with the $m$th ($m=1, \ldots, M$) algorithm running $a_m$ times under randomly and independently chosen settings. The consensus clustering is obtained by applying the average linkage agglomerative clustering on the weighted similarity matrix $\sum_{m=1}^{M}w_m \frac{1}{a_m}\sum_{a=1}^{a_m}\mathbf{B}_{ma}\mathbf{B}_{ma}'$ where $\{\mathbf{B}_{ma}: a=1, \ldots, a_m\}$ are the $a_m$ binary membership indicator matrices for the $m$th algorithm. The weight $w_m$ is defined as $w_m=\frac{a_m}{q_m(1-q_m)}[\sum_{j=1}^{M}\frac{a_j}{q_j(1-q_j)}]^{-1}$, where $q_m$ is the probability of the $m$th algorithm making a correct decision: partition objects from different classes and unite objects from the same class. An estimate of $q_m$ is used to calculate $w_m$. The weight implies that an algorithm producing more stable decisions for a given pair of objects receives higher weight. \cite{Berikov2017} modifies the weighted similarity matrix to be $\sum_{m=1}^{M}w_m \frac{1}{a_m}\sum_{a=1}^{a_m}\gamma_{ma}\mathbf{B}_{ma}\mathbf{B}_{ma}'$, where $\gamma_{ma}$ is the value of a clustering validation index (e.g., the DVI), and $w_m$ is re-defined as
\begin{equation*}
w_m=\frac{\frac{a_m}{\frac{1}{a_m}\sum_{a=1}^{a_m}\gamma_{ma}^2-(\frac{1}{a_m}\sum_{a=1}^{a_m}\gamma_{ma})^2+4(\frac{1}{a_m}\sum_{a=1}^{a_m}\gamma_{ma})^2q_m(1-q_m)}} {\sum_{j=1}^{M}\frac{a_j}{\frac{1}{a_j}\sum_{a=1}^{a_j}\gamma_{ja}^2-(\frac{1}{a_j}\sum_{a=1}^{a_j}\gamma_{ja})^2+4(\frac{1}{a_j}\sum_{a=1}^{a_j}\gamma_{ja})^2q_j(1-q_j)}}.
\end{equation*}
\cite{Berikov2018} further generalizes the framework to fuzzy clustering. Let $\{\mathbf{F}^{ma}: a=1, \ldots, a_m\}$ denote the $a_m$ fuzzy membership matrices for the $m$th algorithm, and $\mathbf{S}^{ma}$ the similarity matrix calculated from the fuzzy matrix $\mathbf{F}^{ma}$: $\mathbf{S}^{ma}_{ij}=\frac{1}{\sqrt{2}}\|\sqrt{\mathbf{F}^{ma}_{i\cdot}}-\sqrt{\mathbf{F}^{ma}_{j\cdot}}\|_2$. A hierarchical agglomerative clustering method is applied on the weighted similarity matrix $\sum_{m=1}^{M}w_m \frac{1}{a_m}\sum_{a=1}^{a_m}\mathbf{S}^{ma}$ to obtain the consensus clustering. Under certain regularity assumptions and the assumption that the data $\mathbb{X}$ are from a mixture distribution, the author gives an analytic expression for the optimal value of $w_m$.

\cite{HUANG2015240} develop two clustering ensemble algorithms. The first algorithm directly applies a hierarchical agglomerative clustering method on the weighted similarity matrix $\mathbf{B}\mathbf{W}\mathbf{B}'$ in which the weight $w_m$ for clustering $C_m$ is the normalized value of $[\frac{\sum_{C\in\mathscr{C}, C\neq C_m}\phi(C_m, C)}{\max_{C_m}\sum_{C\in\mathscr{C}, C\neq C_m}\phi(C_m, C)}]^\alpha$.
$\alpha>0$ is a user-specified parameter to adjust the clustering weights. In the second algorithm, a bipartite graph is constructed in which both objects and clusters are treated as the vertices; an edge links an object and the cluster containing it, or links clusters that have common neighbors. Two clusters are called neighbors if their intersection is non-empty. For an edge linking an object and its cluster, the edge weight is $w_m$, where $C_m$ is the clustering containing the cluster. For an edge linking two clusters $\mathbb{C}^r$ and $\mathbb{C}^v$, the edge weight is proportional to
\begin{equation*}
\sum_{m=1}^{M}(\frac{\sum_{C\in\mathscr{C}, C\neq C_m}\phi(C_m, C)}{\max_{C_m}\sum_{C\in\mathscr{C}, C\neq C_m}\phi(C_m, C)})^\alpha\sum_{l=1}^{k_m}\min\{\frac{|\mathbb{C}_m^l\cap\mathbb{C}^r|}{|\mathbb{C}_m^l\cup\mathbb{C}^r|}, \frac{|\mathbb{C}_m^l\cap\mathbb{C}^v|}{|\mathbb{C}_m^l\cup\mathbb{C}^v|}\},
\end{equation*}
which measures the similarity between clusters $\mathbb{C}^r$ and $\mathbb{C}^v$ w.r.t. their common neighbors and clustering weights. The Tcut algorithm is utilized for partitioning the graph into disjoint sets of vertices. The objects in each disjoint set form a cluster and thus the consensus clustering is obtained.

In \cite{Son2016}, all the base clusterings are fuzzy clusterings. For a fuzzy clustering $C=\{\mathbb{C}^1, \ldots, \mathbb{C}^k\}\in\mathscr{C}$, the fuzzy membership matrix is denoted by $\mathbf{F}\in\mathbb{R}^{n\times k}$, with $\mathbf{F}_{iv}$ being the fuzzy membership of object $\pmb{x}_i$ to cluster $\mathbb{C}^v$: $0\leq\mathbf{F}_{iv}\leq1$ and $\sum_{v=1}^{k}\mathbf{F}_{iv}=1$. The similarity between objects $\pmb{x}_i$ and $\pmb{x}_j$ is defined as $\exp(-\|\mathbf{F}_{i\cdot}-\mathbf{F}_{j\cdot}\|^2_2)$, from which the similarity matrix for clustering $C$ can be determined. The similarity matrix for the ensemble, denoted by $\mathbf{S}$, is a weighted average of the $M$ base-clustering similarity matrices, where the clustering weights are calculated from an internal clustering validation criterion, e.g., the DVI. The fuzzy membership matrix for the consensus clustering, denoted by $\mathbf{F}^*$, is obtained by solving $\min_{\mathbf{F}^*\geq0, \mathbf{F}^*\pmb{1}=\pmb{1}}\|\mathbf{S}-\mathbf{F}^*\mathbf{F}^{*'}\|_F^2.$

Given the ensemble $\mathscr{C}=\{C_1, \ldots, C_M\}$, \cite{rouba2017weighted} calculate the similarity, denoted by $s(C_m, C_t)$, between the clusterings $C_m$ and $C_t$ using the Rand index. Then the $M$ base clusterings are partitioned into several groups by applying a clustering algorithm on the calculated similarity matrix. For each clustering $C_m$, its weight $w_m$ is the normalized value of
\begin{equation*}
\frac{1}{\mbox{size of the group for }C_m}\sum_{C_t\in\mathscr{C}, C_t\neq C_m}s(C_m, C_t)\delta(\mbox{$C_m$ and $C_t$ are in the same group}).
\end{equation*}
Finally, a clustering algorithm is applied on the weighted similarity matrix $\mathbf{B}\mathbf{W}\mathbf{B}'$ to obtain the consensus clustering.

Without discussing how the clustering weights are determined, \cite{8013756} develop an iterative algorithm for solving problem (\ref{weighted}) in which $\{C_1, \ldots, C_M\}$ is an ensemble of fuzzy clusterings. Let $\pmb{F}$ and $\pmb{F}^m$ respectively denote the fuzzy membership matrices for clusterings $C=\{\mathbb{C}^1, \ldots, \mathbb{C}^k\}$ and $C_m=\{\mathbb{C}_m^1, \ldots, \mathbb{C}_m^{k_m}\}$. The similarity between clusters $\mathbb{C}^r$ and $\mathbb{C}_m^v$ ($r=1, \ldots, k$ and $v=1, \ldots, k_m$) is defined as $s_{rv}^m=\langle(\pmb{F}_{\cdot r})^\alpha, \pmb{F}^m_{\cdot v}\rangle$, where $\alpha (>1)$ is a user-specified fuzzy factor. For $r=1, \ldots, k$, define $s_{r+}^m=\sum_{v=1}^{k_m}s_{rv}^m=\|(\pmb{F}_{\cdot r})^\alpha\|_1$ and $\pmb{p}_r^m=(\frac{s_{r1}^m}{s_{r+}^m}, \ldots, \frac{s_{rk_m}^m}{s_{r+}^m})$. A highly biased $\pmb{p}_r^m$ implies a heavy overlap between the $r$th cluster of the consensus clustering $C$ and certain cluster in $C_m$. The similarity function in (\ref{weighted}) is defined as $\phi(C, C_m)=\sum_{r=1}^{k}\|(\pmb{F}_{\cdot r})^\alpha\|_1\times g(\pmb{p}_r^m)-\sum_{i=1}^{n}\|(\pmb{F}_{i\cdot})^\alpha\|_1\times g(\pmb{F}^m_{i\cdot})$, where $g(\cdot)$ is a convex function and hence favors biased $\pmb{p}_r^m$. A fuzzy c-means type algorithm is developed to obtain the consensus fuzzy clustering $C$.

To detect protein complexes, \cite{WuMin2017} modify the co-occurrence matrix by incorporating available co-complex information between proteins: $[\mathbf{B}_m\mathbf{B}_m']_{ij}\leftarrow\frac{1+\tau_{ij}}{2}\delta([\mathbf{B}_m\mathbf{B}_m']_{ij}=1)$. $\tau_{ij}$ is a co-complex affinity score between proteins $\pmb{x}_i$ and $\pmb{x}_j$. With an initial set of clustering weights $\{w_1, \ldots, w_M\}$, the clustering $C=\{\mathbb{C}^1, \ldots, \mathbb{C}^k\}$ obtained by applying a hierarchical clustering method on the weighted similarity matrix $\sum_{m=1}^{M}w_m\mathbf{B}_m\mathbf{B}_m'$ is set as the reference clustering. Then the weight $w_m$ for clustering $C_m$ is updated: $w_m\leftarrow\frac{\sum_{v=1}^{k_m}\max_{r=1, \ldots, k}|\mathbb{C}^r\cap\mathbb{C}_m^v|}{\sum_{v=1}^{k_m}|\cup_{r=1}^k(\mathbb{C}^r\cap\mathbb{C}_m^v)|}$.
$w_m$ measures the proportion of the clusters in $C_m$ that are matched by the clusters in the reference clustering $C$. Repeat the following two steps until the Pearson correlation between the updated weights and the weights in the preceding round is larger than the threshold 0.9: (1) update the reference clustering by applying the hierarchical clustering method on $\sum_{m=1}^{M}w_m\mathbf{B}_m\mathbf{B}_m'$; (2) update the weights via the previous equation. The final reference clustering is the consensus clustering.

The framework proposed by \cite{Yousefnezhad2018} addresses three aspects of weighted clustering ensemble: independency, decentralization and diversity. For independency, a linear transformation is performed on the data $\mathbb{X}$ such that the transformed data $\mathbb{Y}$ have the lowest correlations between its features. For decentralization, prior knowledge on must-links and cannot-links is utilized. In particular, a projection matrix is determined such that, in the projected low-dimensional data $\mathbb{Z}$, objects that must link are close and objects that cannot link are far from each other. A clustering ensemble is obtained by applying different clustering algorithms on the projected data $\mathbb{Z}$. For diversity, a metric is developed for evaluating the diversity of clustering $C_m$:
\begin{equation*}
1-\frac{2\times\min_{\mathbb{C}_m^v\in C_m}|\mathbb{C}_m^v|\log(\frac{|\mathbb{C}_m^v|}{n})}{\max_{\mathbb{C}_m^v\in C_m}|\mathbb{C}_m^v|\log(\frac{|\mathbb{C}_m^v|}{n})+\max_{C_m\in\mathscr{C}}\max_{\mathbb{C}_m^v\in C_m}|\mathbb{C}_m^v|\log(\frac{|\mathbb{C}_m^v|}{n})}.
\end{equation*}
The above metric evaluates the diversity of $C_m$ w.r.t. the other clusterings, and a small value represents low diversity. The consensus clustering is obtained by applying the average linkage agglomerative clustering on the weighted similarity matrix $\mathbf{B}\mathbf{W}\mathbf{B}'$, in which the weight of a clustering is calculated from its diversity.

To deal with high-dimensional data and incorporate prior knowledge into the clustering results, \cite{8323237} propose to sub-sample the feature space and, given a subset of features, select relatively important constraints from the original constraint set. Given a subset of features and the corresponding selected constraints, a semi-supervised clustering algorithm is employed to construct the base clustering. The consensus clustering is obtained as follows. Firstly, an initial set of $2M$ clustering weight vectors is generated. By applying the Ncut algorithm on the resulted $2M$ weighted similarity matrices, we obtain $2M$ candidate consensus clusterings. To compare different clusterings, a performance criterion is defined, which is the sum of the within-cluster variance and the number of unsatisfied constraints. Then the $2M$ clustering weight vectors are modified via uniform competition; that is, given two clustering weight vectors, their $l$th elements will be exchanged if a uniform random variable takes a value larger than 0.5, $l=1, \ldots, M$. Apply the Ncut algorithm on the new $2M$ weighted similarity matrices, and compare the two best-performance candidate consensus clusterings respectively in the current and preceding iterations. Stop the iteration if the best-performance candidate consensus clustering stays unchanged for a few times.

\subsection{Weighting Clusters}\label{FWC}
\cite{YANG2016} bring the weighted clustering ensemble paradigm into the problem of community detection where communities (i.e., clusters) in each partition (i.e., clustering) can overlap: $\mathbb{C}_m^r\cap\mathbb{C}_m^v\neq\varnothing$, $1\leq r<v\leq k_m$. They first assign a weight to each cluster:
\begin{equation*}
w_{\mathbb{C}_m^r}=1+\frac{1}{|E_{i}(\mathbb{C}_m^r)|+|E_{o}(\mathbb{C}_m^r)|}\sum_{(\pmb{x}_i,\pmb{x}_j)\in E(\mathbb{C}_m^r)}\{[\frac{1}{M}\mathbf{B}\mathbf{B}']_{ij}\log_2([\frac{1}{M}\mathbf{B}\mathbf{B}']_{ij}) +(1-[\frac{1}{M}\mathbf{B}\mathbf{B}']_{ij})\log_2(1-[\frac{1}{M}\mathbf{B}\mathbf{B}']_{ij})\},
\end{equation*}
where $E(\mathbb{C}_m^r)=E_{i}(\mathbb{C}_m^r)\cup E_{o}(\mathbb{C}_m^r)$, and $E_{i}(\mathbb{C}_m^r)$ and $E_{o}(\mathbb{C}_m^r)$ are two sets of edges: an edge $(\pmb{x}_i,\pmb{x}_j)\in E_{i}(\mathbb{C}_m^r)$ if $\pmb{x}_i\in\mathbb{C}_m^r$ and $\pmb{x}_j\in\mathbb{C}_m^r$; an edge $(\pmb{x}_i,\pmb{x}_j)\in E_{0}(\mathbb{C}_m^r)$ if either $\pmb{x}_i\in\mathbb{C}_m^r$ or $\pmb{x}_j\in\mathbb{C}_m^r$. If all edges in $E_{i}(\mathbb{C}_m^r)$ have $[\frac{1}{M}\mathbf{B}\mathbf{B}']_{ij}=1$ and all edges in $E_{o}(\mathbb{C}_m^r)$ have $[\frac{1}{M}\mathbf{B}\mathbf{B}']_{ij}=0$, then $w_{\mathbb{C}_m^r}=1$, implying that cluster $\mathbb{C}_m^r$ is stable. Otherwise, if all edges in $E(\mathbb{C}_m^r)$ have $[\frac{1}{M}\mathbf{B}\mathbf{B}']_{ij}=0.5$, then $w_{\mathbb{C}_m^r}=0$, implying that cluster $\mathbb{C}_m^r$ is unstable. The authors then define a membership matrix $\mathbf{F}^m$ for clustering $C_m$:
\begin{equation*}
\mathbf{F}^m_{ir}=\frac{\delta(\pmb{x}_i\in\mathbb{C}_m^r)}{\mbox{total number of clusters in $C_m$ that contain $\pmb{x}_i$}},
\end{equation*}
and from which a new similarity matrix $\mathbf{S}$: $\mathbf{S}_{ij}=\frac{1}{M}\sum_{m=1}^{M}\sum_{r=1}^{k_m}g(\mathbf{F}^m_{ir}, \mathbf{F}^m_{jr})\times w_{\mathbb{C}_m^r}$, where $g(\cdot, \cdot)$ is a suitable fuzzy t-norm. To reduce the influence of noise and improve the algorithm speed, the similarity matrix is further filtered: $\mathbf{S}_{ij}\leftarrow\mathbf{S}_{ij}\times\delta(\mathbf{S}_{ij}> \mbox{a random value from [0, 1]}).$
Repeat the following two steps until all base clusterings in an ensemble are the same: (1) apply a non-deterministic overlapping community detection algorithm on $\mathbf{S}$ for $M$ times to generate a new ensemble of $M$ base clusterings; (2) update the cluster weights and the similarity matrix $\mathbf{S}$ from the new ensemble. The final unique base clustering is the desired consensus clustering.

\cite{Huang2017} construct a graph in which all the clusters in the ensemble $\mathscr{C}$ are treated as the vertices, and the weight for an edge linking two clusters is their Jaccard coefficient. The Ncut algorithm is adopted to partition the graph into a certain number of meta-clusters $\{\tilde{\mathbb{C}}^1, \tilde{\mathbb{C}}^2, \ldots\}$ with each meta-cluster being a set of clusters (vertices). Then an object $\pmb{x}_i$ will be assigned to a meta-cluster $\tilde{\mathbb{C}}^k$ having the highest weighted vote $\frac{1}{|\tilde{\mathbb{C}}^k|}\sum_{\mathbb{C}\in\tilde{\mathbb{C}}^k}w_{\mathbb{C}}\delta(\pmb{x}_i\in \mathbb{C})$, where the weight for a cluster $\mathbb{C}$ is $w_{\mathbb{C}}=\exp(-\frac{1}{\alpha M}\sum_{m=1}^M H(\frac{|\mathbb{C}\cap \mathbb{C}_m^1|}{|\mathbb{C}|}, \ldots, \frac{|\mathbb{C}\cap \mathbb{C}_m^{k_m}|}{|\mathbb{C}|}))$, which implies the uncertainty of a cluster w.r.t. all the clusters in the ensemble.
Using the above definition of cluster weight, \cite{7932479} further propose another two algorithms. In the first algorithm, a hierarchical agglomerative clustering method is applied on a similarity matrix $\mathbf{S}$: $\mathbf{S}_{ij}=\frac{1}{M}\sum_{m=1}^M\sum_{r=1}^{k_m}w_{\mathbb{C}_m^r}\delta(\pmb{x}_i\in \mathbb{C}_m^r)[\mathbf{B}_m\mathbf{B}_m']_{ij}$. The intuition is that objects that co-occur in more reliable clusters (with higher weights) are more likely to belong to the same cluster in the true clustering. In the second algorithm, a bipartite graph is constructed: both objects and clusters are treated as the vertices; an edge only links an object and the cluster containing it; the weight of an edge is the weight of the linked cluster. The Tcut algorithm is utilized for partitioning the graph into disjoint sets of vertices. The objects in each disjoint set form a cluster, and thus the consensus clustering is obtained.
According to the edge weights of the graph in \cite{Huang2017}, \cite{8525437} construct a one-step transition probability matrix $\pmb{P}^{(1)}$ for the random walk process on the graph. The $r$th row $\pmb{P}^{(1)}_{r\cdot}$ is a probability distribution for transiting from cluster/node $\mathbb{C}^r$ to any other cluster. A new similarity between clusters $\mathbb{C}^r$ and $\mathbb{C}^v$ is defined by calculating the cosine similarity between $\{\pmb{P}^{(1)}_{r\cdot}, \pmb{P}^{(2)}_{r\cdot}, \ldots, \pmb{P}^{(t)}_{r\cdot}\}$ and $\{\pmb{P}^{(1)}_{v\cdot}, \pmb{P}^{(2)}_{v\cdot}, \ldots, \pmb{P}^{(t)}_{v\cdot}\}$, where $\pmb{P}^{(t)}=\underbrace{\pmb{P}^{(1)} \cdots \pmb{P}^{(1)}}\limits_{t}$ is the $t$-step transition probability matrix. After replacing each edge weight with the new cluster similarity, the approach in \cite{Huang2017} is taken to obtain the consensus clustering.

The certainty that a given ensemble has about a cluster is considered by \cite{Nazari2017} as the reliability of that cluster. More precisely, for any cluster $\mathbb{C}$ and clustering $C_m\in\mathscr{C}$, let $P_{C_m}(\mathbb{C})$ denote the projection of $\mathbb{C}$ on $C_m$: $P_{C_m}(\mathbb{C})=\{\mathbb{C}\cap\mathbb{C}_m^1, \ldots, \mathbb{C}\cap\mathbb{C}_m^{k_m}\}$. $P_{C_m}(\mathbb{C})$ can be treated as a clustering of the objects in $\mathbb{C}$, and hence $P_{\mathscr{C}}(\mathbb{C})$ can be treated as a clustering ensemble for the objects in $\mathbb{C}$. We might let $\pmb{B}_m^\mathbb{C}$ denote the binary membership indicator matrix for $P_{C_m}(\mathbb{C})$. The reliability of $\mathbb{C}$ w.r.t. the ensemble $\mathscr{C}$ is
\begin{equation*}
\mbox{reliability}(\mathbb{C}, \mathscr{C})=\frac{1}{|\mathbb{C}|(|\mathbb{C}|-1)}\sum_{i,j=1}^{|\mathbb{C}|} [\frac{1}{M}\sum_{m=1}^{M}\pmb{B}_m^\mathbb{C}{\pmb{B}_m^{\mathbb{C}}}']_{ij}-\frac{1}{|\mathbb{C}|-1}.
\end{equation*}
Define $\pmb{S}$ as the cluster-level weighted similarity matrix for the ensemble $\mathscr{C}$:
\begin{equation*}
\pmb{S}_{ij}=\frac{\sum_{m=1}^{M}\sum_{v=1}^{k_m}\mbox{reliability}(\mathbb{C}_m^v, \mathscr{C})\times\delta(\pmb{x}_i\in\mathbb{C}_m^v )\delta(\pmb{x}_j\in\mathbb{C}_m^v )\delta(\mbox{reliability}(\mathbb{C}_m^v, \mathscr{C})>\mbox{MR})}{\sum_{m=1}^{M}\sum_{v=1}^{k_m}\delta(\pmb{x}_i\in\mathbb{C}_m^v )\delta(\pmb{x}_j\in\mathbb{C}_m^v )\delta(\mbox{reliability}(\mathbb{C}_m^v, \mathscr{C})>\mbox{MR})},
\end{equation*}
where MR is the median of all the cluster reliabilities. The authors propose two approaches to obtain the consensus clustering. In the first approach, the consensus clustering is obtained by applying the average linkage agglomerative clustering on the similarity matrix $\pmb{S}$. In the second approach, a bipartite graph is constructed in which objects and clusters with reliabilities larger than MR are treated as the vertices, and an edge only links an object and the cluster containing it. The weight of an edge is the reliability of the linked cluster. A graph partitioning algorithm is applied on the bipartite graph to obtain the consensus clustering.
\cite{Bagherinia20211} extend the reliability idea to fuzzy clustering, where the unreliability of a fuzzy cluster $\mathbb{C}$ with respect to a fuzzy base clustering $C_m$, denoted by ``$\mbox{unreliability}(\mathbb{C}, C_m)$'', is defined as the entropy of the similarities between $\mathbb{C}$ and all the clusters in $C_m$. The unreliability of $\mathbb{C}$ with respect to the ensemble is simply the sum: $\mbox{unreliability}(\mathbb{C}, \mathscr{C})=\sum_{m=1}^{M}\mbox{unreliability}(\mathbb{C}, C_m)$, and the counterpart ``$\mbox{reliability}(\mathbb{C}, \mathscr{C})$'' is obtained by applying a transformation on $\mbox{unreliability}(\mathbb{C}, \mathscr{C})$. The weighted similarity matrix $\pmb{S}$ for the fuzzy ensemble is defined as:
$$\pmb{S}_{ij}=\frac{1}{M}\sum_{m=1}^{M}\sum_{v=1}^{k_m}\mathbf{F}^m_{iv}\times\mathbf{F}^m_{jv}\times \mbox{reliability}(\mathbb{C}_m^v, \mathscr{C}),$$
where $\mathbf{F}^m$ is the membership matrix for clustering $C_m$. The two approaches proposed in \cite{Nazari2017} are followed to obtain the consensus clustering.
In \cite{doi:10.1080/09540091}, the cluster-level weight is defined as
\begin{equation*}
w(\mathbb{C}_r^k)=\exp(\frac{1}{M}\sum_{m=1, m\neq r}^{M}\sum_{v=1}^{k_m}\frac{|\mathbb{C}_m^v|}{n}\log(\frac{|\mathbb{C}_r^k\cap\mathbb{C}_m^v|}{|\mathbb{C}_m^v|})).
\end{equation*}
The cluster-level weighted similarity matrix $\pmb{S}$ for the ensemble is defined as
\begin{equation*}
\pmb{S}_{ij}=\frac{1}{M}\sum_{m=1}^{M}\sum_{v=1}^{k_m}w(\mathbb{C}_m^v)\delta(\pmb{x}_i\in\mathbb{C}_m^v)\delta(\pmb{x}_j\in\mathbb{C}_m^v).
\end{equation*}
The consensus clustering is obtained by applying the average linkage agglomerative clustering on the weighted similarity matrix $\pmb{S}$. The work also suggests an ensemble selection method.

\subsection{Weighting Objects}
Based on the similarity matrix $\frac{1}{M}\mathbf{B}\mathbf{B}'$, \cite{Hanan2003} identify a set of nearest neighbors for each object $\pmb{x}_i$: $\mathscr{N}_i=\{\pmb{x}_j: [\frac{1}{M}\mathbf{B}\mathbf{B}']_{ij}>$ a predetermined threshold$\}$. They then construct a graph in which the vertices are the objects; an edge exists between vertices $\pmb{x}_i$ and $\pmb{x}_j$ if $\pmb{x}_i$ is a $k$-nearest neighbor of $\pmb{x}_j$, or vice versa. Define $\mathscr{N}_{ij}=\mathscr{N}_i\cap \mathscr{N}_j$. The weight $w_i$ of each vertex (object) is computed as $w_i=\frac{\mbox{a balancing factor}}{\sum\limits_{\pmb{x}_j\in \mathscr{N}_i}\sum\limits_{\pmb{x}_l\in \mathscr{N}_{ij}}[\frac{1}{M}\mathbf{B}\mathbf{B}']_{il}\times[\frac{1}{M}\mathbf{B}\mathbf{B}']_{jl}}$. The weights assigned to the vertices are meant to reflect in the graph the varying cluster sizes. The weight of the edge linking vertices $\pmb{x}_i$ and $\pmb{x}_j$, if exists, is computed as
\begin{equation*}
\frac{\sum\limits_{\pmb{x}_l\in \mathscr{N}_{ij}}[\frac{1}{M}\mathbf{B}\mathbf{B}']_{il}\times[\frac{1}{M}\mathbf{B}\mathbf{B}']_{jl}}{\sum\limits_{\pmb{x}_l\in \mathscr{N}_{ij}}[\frac{1}{M}\mathbf{B}\mathbf{B}']_{il}^2+\sum\limits_{\pmb{x}_l\in \mathscr{N}_{ij}}[\frac{1}{M}\mathbf{B}\mathbf{B}']_{jl}^2 -\sum\limits_{\pmb{x}_l\in \mathscr{N}_{ij}}[\frac{1}{M}\mathbf{B}\mathbf{B}']_{il}\times[\frac{1}{M}\mathbf{B}\mathbf{B}']_{jl}}.
\end{equation*}
After its construction, the graph is then partitioned using the graph partitioning package METIS. The objective of the partitioning algorithm is to minimize the edge-cut, subject to the constraint that the weights of the vertices are equally distributed among the clusters.

\cite{Topchy2004} devise an adaptive data sampling scheme that focuses on uncertain and problematic objects. In the $m$th iteration, given a set of object weights $\{w_1, \ldots, w_n\}$ calculated from the preceding iteration, a subset of objects is generated via weighted sampling, and the k-means algorithm is applied on the subset to obtain the base clustering $C_m$. The Hungarian algorithm for minimal weight bipartite matching problem is used to re-label $C_m$. After re-labeling, the consistency index of object $\pmb{x}_i$ can be calculated:
\begin{equation*}
\mbox{consistency}(\pmb{x}_i)=\max_{r=1, \ldots, k}\frac{\mbox{number of times $\pmb{x}_i$ belongs to the $r$th cluster}}{\mbox{number of times $\pmb{x}_i$ is sampled}}.
\end{equation*}
The weight/sampling probability for object $\pmb{x}_i$ is then updated: $w_i\leftarrow\frac{\alpha w_i+1-\mbox{consistency}(\pmb{x}_i)}{\sum_{j=1}^{n}[\alpha w_j+1-\mbox{consistency}(\pmb{x}_j)]}$. \cite{PARVIN20131433} instead calculate the uncertainty value of object $\pmb{x}_i$:
\begin{equation*}
\mbox{uncertainty}(\pmb{x}_i)=-H([\frac{1}{m}\sum_{t=1}^{m}\mathbf{B}_t\mathbf{B}_t']_{i\cdot}),
\end{equation*}
and update the sampling probability for object $\pmb{x}_i$ with $\alpha w_i+(1-\alpha)\frac{w_i\times\mbox{uncertainty}(\pmb{x}_i)}{\sum_{j=1}^{n}w_j\times\mbox{uncertainty}(\pmb{x}_j)}$. The consensus clustering is obtained by applying an existing clustering ensemble technique.

\cite{FROSSYNIOTIS2004641} develop an iterative clustering algorithm, called boost-clustering. In the first iteration, clustering $C_1$ is obtained by applying an arbitrary clustering algorithm on a bootstrap replicate of $\mathbb{X}$ via weighted sampling; every object has the (sampling) weight $w_i=1/n$. From $C_1$, we can determine a membership matrix $\mathbf{F}^1$ in which $\mathbf{F}^1_{ir}$ represents the membership degree of the $i$th object to the $r$th cluster. For example, the membership degree $\mathbf{F}^1_{ir}$ can be calculated from the distance between the $i$th object and the centroid of the $r$th cluster. The pseudoloss in the first iteration is $\epsilon_1=\frac{1}{2}\sum_{i=1}^{n}w_i(1-\max_r\mathbf{F}^1_{ir}+\min_r\mathbf{F}^1_{ir})$ or $\epsilon_1=\frac{1}{2}\sum_{i=1}^{n}w_i H(\mathbf{F}^1_{i\cdot})$. Then the object weight $w_i$ is updated by the normalized value of $w_i(\frac{1-\epsilon_1}{\epsilon_1})^{1-\max_r\mathbf{F}^1_{ir}+\min_r\mathbf{F}^1_{ir}}$ or $w_i(\frac{1-\epsilon_1}{\epsilon_1})^{H(\mathbf{F}^1_{i\cdot})}$. Hence, the higher the weight $w_i$, the more difficult it is for $\pmb{x}_i$ to be clustered. In the second iteration, $C_2$ is obtained by clustering the second bootstrap data generated by weighted sampling using the updated object weights. Then the clusters in $C_2$ are relabelled by solving the labeling correspondence problem with reference to $C_1$. After cluster correspondence, the second membership matrix $\mathbf{F}^2$ is obtained. From $\mathbf{F}^2$, the second pseudoloss $\epsilon_2$ is obtained, and object weights can be updated. $C_1$ and $C_2$ are combined into an aggregated clustering in which the cluster for object $\pmb{x}_i$ is
\begin{equation}
 \arg\max_{r=1, \ldots, k}\sum_{v=1}^{\mbox{c.i.n.}} \frac{\log(\frac{1-\epsilon_v}{\epsilon_v})}{\sum_{l=1}^{\mbox{c.i.n.}}\log(\frac{1-\epsilon_l}{\epsilon_l})}\mathbf{F}^v_{ir},
 \label{strange}
\end{equation}
where c.i.n. is short for ``current iteration number'' (e.g., in the second iteration, c.i.n.=2). The aggregated clustering will be used as the reference in the next round labeling correspondence problem, and the final aggregated clustering is the consensus clustering. In addition to weighting objects, Equation (\ref{strange}) implies that the consensus clustering is obtained by weighted voting, where the weight for each clustering is a measure of its quality.
The idea of pseudoloss has been adopted by other works. \cite{Doan:2011:SIC} apply the Gaussian mixture model on the boosting subsets to generate the base clusterings. The only difference is that, given the object weights $\{w_1, w_2, \ldots, w_n\}$, the boosting subset is composed of $\exp(H(w_1, \ldots, w_n))$ (round to the nearest integer) data points having the highest weights. To reduce the learning sensitivity w.r.t. noise and outliers, \cite{YangJiang2016} propose a hybrid sampling scheme: in the $m$th iteration, a candidate subset is generated by random sampling with replacement from $\mathbb{X}$, and the subset for building clustering $C_m$ is obtained by weighted sampling from the candidate subset. Given clustering $C_m$, the clustering weight $\textsf{w}_m$ is defined as
\begin{equation*}
\textsf{w}_m=\frac{\exp(-\frac{2}{n}\sum_{i=1}^{n}[1-\max_r\mathbf{F}^m_{ir}+\min_r\mathbf{F}^m_{ir}])} {\sum_{l=1}^{m}\exp(-\frac{2}{n}\sum_{i=1}^{n}[1-\max_r\mathbf{F}^l_{ir}+\min_r\mathbf{F}^l_{ir}])},
\end{equation*}
and the sampling weight for object $\pmb{x}_i$ in the $(m+1)$st iteration is updated by
\begin{equation*}
w_i\leftarrow\frac{w_i\exp(\textsf{w}_m[1-\max_r\mathbf{F}^m_{ir}+\min_r\mathbf{F}^m_{ir}])}{\sum_{j=1}^{n}w_j\exp(\textsf{w}_m[1-\max_r\mathbf{F}^m_{jr}+\min_r\mathbf{F}^m_{jr}])}.
\end{equation*}
The consensus clustering is obtained by applying a conventional clustering algorithm on the concatenated matrix $[\sqrt{\textsf{w}_1}\mathbf{F}^1, \sqrt{\textsf{w}_2}\mathbf{F}^2, \ldots, \sqrt{\textsf{w}_M}\mathbf{F}^M]$.

\cite{4406080} combine the boosting framework with fuzzy clustering. In the $m$th boosting iteration, the fuzzy membership matrix $\mathbf{F}^m$ is obtained, and the similarity matrix $\mathbf{S}$ is updated: $\mathbf{S}_{ij}\leftarrow\alpha\mathbf{S}_{ij}+(1-\alpha)\frac{\langle\mathbf{F}^m_{i\cdot},~ \mathbf{F}^m_{j\cdot}\rangle}{\|\mathbf{F}^m_{i\cdot}\|_2^2+\|\mathbf{F}^m_{j\cdot}\|_2^2-\langle\mathbf{F}^m_{i\cdot},~ \mathbf{F}^m_{j\cdot}\rangle}$. Then the object weight $w_i$ is updated to be $1+\frac{1}{\log(k)}H(\mathbf{F}^m_{i\cdot})[1+\frac{1}{\log(k)}H(\mathbf{F}^m_{i\cdot})]$. The CSPA is applied on the final similarity matrix $\mathbf{S}$ to obtain the consensus clustering.
\cite{Saffari2008} combine the boosting framework with model-based clustering. In the $m$th boosting iteration, the base clustering $C_m$ is obtaining by minimizing the loss $\frac{1}{n}\sum_{i=1}^{n}w_i\sum_{r=1}^{k}\Pr(\pmb{x}_i \mbox{ belongs to the $r$th cluster}|C_m)\ell(\pmb{x}_i, r|C_m)$, where $\ell(\pmb{x}_i, r|C_m)$ is the loss for assigning object $\pmb{x}_i$ to the $r$th cluster. Then the object weight $w_i$ is updated to be $w_i\exp(\sum_{r=1}^{k}\Pr(\pmb{x}_i \mbox{ belongs to the $r$th cluster}|C_m)\ell(\pmb{x}_i, r|C_m))$. After solving the labeling correspondence problem, a consensus function learning method is applied on the ensemble to derive the final clusters.

\cite{RASHEDI201383} propose a boosting-based hierarchical clustering ensemble technique. In the first iteration, a subset of objects is randomly generated (without replacement) from $\mathbb{X}$. An arbitrary hierarchical clustering method is applied on the subset to create the first hierarchical clustering $C_1$. For each object, there exist a Euclidean-distance vector (with each entry being the Euclidean distance between the object and any other object in $\mathbb{X}$) and a hierarchical-distance vector (with each entry being the dissimilarity, e.g. the cophenetic difference, between the object and any other object w.r.t. the hierarchical clustering $C_1$). The weight of an object is calculated from the correlation coefficient between its Euclidean-distance vector and hierarchical-distance vector, and a higher modulus correlation coefficient gives a lower weight. After assigning weights to objects, in the second iteration, a subset of objects is generated from $\mathbb{X}$ by weighted sampling such that an object with a low modulus correlation coefficient gets a high probability of being selected. The hierarchical clustering method is applied on the second subset to create the second hierarchical clustering $C_2$. $C_1$ and $C_2$ are then aggregated by certain combination method to obtain an aggregated clustering, and the aggregated clustering is used to update the hierarchical-distance vectors and hence the object weights. Repeat the weighted sampling, clustering aggregation and weights updating steps for many times. The final aggregated clustering is treated as the consensus clustering.

In \cite{6729547} and \cite{Ren2017}, given the ensemble $\mathscr{C}$, the similarity matrix $\frac{1}{M}\mathbf{B}\mathbf{B}'$ is used to quantify the level of uncertainty in clustering two objects $\pmb{x}_i$ and $\pmb{x}_j$: $[\frac{1}{M}\mathbf{B}\mathbf{B}']_{ij}(1-[\frac{1}{M}\mathbf{B}\mathbf{B}']_{ij})$. Then the weight for object $\pmb{x}_i$ is $w_i=\frac{4}{n}\sum_{j=1}^{n} [\frac{1}{M}\mathbf{B}\mathbf{B}']_{ij}(1-[\frac{1}{M}\mathbf{B}\mathbf{B}']_{ij})$, which indicates how hard it is to cluster object $\pmb{x}_i$. To incorporate the object weights into the consensus clustering, three different algorithms are developed. The first algorithm modifies the Meta-CLustering Algorithm \citep{Strehl:2003} in that the edge connecting two clusters (vertices) $\mathbb{C}^r$ and $\mathbb{C}^v$ is assigned the weight $\frac{\sum_{\pmb{x}_i\in\mathbb{C}^r\cap\mathbb{C}^v}w_i}{\sum_{\pmb{x}_i\in\mathbb{C}^r\cup\mathbb{C}^v}w_i}$. The second algorithm moves the centroid of each cluster towards the objects that are hard to cluster. Specifically, for a base clustering $C=\{\mathbb{C}^1, \ldots, \mathbb{C}^k\}\in\mathscr{C}$ with centroids $\{\pmb{c}^1, \ldots, \pmb{c}^k\}$, its weighted centroids are $\tilde{\pmb{c}}^r=\frac{\sum_{\pmb{x}_i\in \mathbb{C}^r} w_i\pmb{x}_i}{\sum_{\pmb{x}_i\in \mathbb{C}^r}w_i},$ $r=1, \ldots, k$. The similarity between an object and a weighted centroid is defined as $\exp(-\frac{1}{\alpha}\|\pmb{x}_i-\tilde{\pmb{c}}^r\|_2^2)$, based on which a probability mass function w.r.t. clustering $C$ can be obtained for every object: $(\Pr(\pmb{x}_i\in\mathbb{C}^1), \ldots, \Pr(\pmb{x}_i\in\mathbb{C}^k))$. The following steps are the same as those in \cite{Muna2006} (see Section \ref{VWF}): the similarity between objects $\pmb{x}_i$ and $\pmb{x}_j$ w.r.t. clustering $C_m$ is the cosine similarity between their probability mass functions $(\Pr(\pmb{x}_i\in\mathbb{C}^1_m), \ldots, \Pr(\pmb{x}_i\in\mathbb{C}^{k_m}_m))$ and $(\Pr(\pmb{x}_j\in\mathbb{C}^1_m), \ldots, \Pr(\pmb{x}_j\in\mathbb{C}^{k_m}_m))$; then the similarity matrix $\mathbf{S}_m$ for clustering $C_m$ is determined; METIS is applied on the graph $G(\frac{1}{M}\sum_{m=1}^M \mathbf{S}_m)$ to obtain the consensus clustering. In the third algorithm, a bipartite graph is constructed in which both objects and clusters are treated as the vertices. An edge only links an object (say, $\pmb{x}_i$) and a cluster (say, $\mathbb{C}_m^r$), and the edge weight is the probability mass $\Pr(\pmb{x}_i\in\mathbb{C}_m^r)$. METIS is applied on the bipartite graph, so that the edge weight-cut is minimized. The objects in each disjoint set form a cluster, and thus the consensus clustering is obtained.

Given the current base clusterings $\{C_1, \ldots, C_m\}$, \cite{10.1007} apply a consensus clustering algorithm to the similarity matrix $\frac{1}{m}\sum_{t=1}^{m}\mathbf{B}_t\mathbf{B}_t'$ to obtain the current consensus clustering $\{\mathbb{C}^1, \ldots, \mathbb{C}^{k}\}$. The degree of confidence of assigning object $\pmb{x}_i$ to its cluster $\mathbb{C}^r\in\{\mathbb{C}^1, \ldots, \mathbb{C}^{k}\}$ is
\begin{equation*}
\mbox{conf}(\pmb{x}_i)=\frac{1}{|\mathbb{C}^r|-1}\sum_{\pmb{x}_j\in\mathbb{C}^r\setminus\{\pmb{x}_i\}} [\frac{1}{m}\sum_{t=1}^{m}\mathbf{B}_t\mathbf{B}_t']_{ij}-\max_{v:~ v\neq r}\frac{1}{\mathbb{C}^v}\sum_{\pmb{x}_j\in\mathbb{C}^v}[\frac{1}{m}\sum_{t=1}^{m}\mathbf{B}_t\mathbf{B}_t']_{ij}.
\end{equation*}
If the average similarity of $\pmb{x}_i$ w.r.t. the other objects in $\mathbb{C}^r$ is higher than the average similarity to the objects in any other cluster, then the confidence for assigning $\pmb{x}_i$ to $\mathbb{C}^r$ is high. Given the confidence levels, one can calculate different types of object weights. For example, if we want to focus on objects with high degrees of confidence, then the weight for object $\pmb{x}_i$ can be the normalized value of $\mbox{conf}(\pmb{x}_i)$. Given the object weights, the $(m+1)$st base clustering is obtained by applying a weighted $k$-means algorithm on $\mathbb{X}$, in which a cluster centroid is updated by the weighted average of the cluster members. The consensus clustering is obtained by applying the consensus clustering algorithm on the final similarity matrix $\frac{1}{M}\sum_{m=1}^{M}\mathbf{B}_m\mathbf{B}_m'$.

\cite{Liu2015SEC} show that applying the normalized cuts spectral clustering on $\mathbf{B}\mathbf{B}'$ (to obtain the consensus clustering) has the equivalent objective function to the weighted $k$-means clustering on data $\{\mathbf{B}_{i\cdot}: i=1, \ldots, n\}$, where the $i$th data point $\mathbf{B}_{i\cdot}$ is weighted by $\frac{1}{\langle\mathbf{1},~[\mathbf{B}\mathbf{B}']_{i\cdot}\rangle}$. $\mathbf{1}$ is a column vector of 1's. The equivalence to the weighted $k$-means clustering dramatically decreases the time and space complexity of the algorithm.

\subsection{Weighting Features}
In DNA microarray data clustering, the data set $\mathbb{X}$ represents $n$ samples of $p$ genes. It is usually the case that the number of features far exceeds the number of samples by many orders of magnitude. To cluster the $n$ samples, \cite{Amaratunga2008} perform weighted sub-sampling on the $p$ features (genes). The weight $w_r$ ($r=1, \ldots, p$) for the $r$th feature is $w_r=(\mbox{rank of the variance of the $r$th gene}+\alpha)^{-1}$, where $\alpha$ is such that $1\%$ of the genes with the highest variance have a combined probability of $20\%$ of being selected. The feature weight formulation implies that a feature with a high variance will be given a low weight. To construct clustering $C_m$, first select $\sqrt{p}$ genes via weighted sampling without replacement using the weights $\{w_r\}_{r=1}^p$; then randomly select $n$ samples with replacement, discarding replicate samples; finally, run a clustering algorithm on the resulting subset to obtain $C_m$ with $\sqrt{n}$ clusters. After obtaining the ensemble, calculate the similarity matrix $\mathbf{S}$:
\begin{equation*}
\mathbf{S}_{ij}=\frac{[\mathbf{B}\mathbf{B}']_{ij}}{\mbox{number of subsets that include both $\pmb{x}_i$ and $\pmb{x}_j$}}.
\end{equation*}
The similarity matrix is the input to a hierarchical clustering method to obtain the consensus clustering.

Following \cite{FROSSYNIOTIS2004641}, instead of weighting the objects, \cite{5302706} proposes to weight the features. In the $(m+1)$st boosting iteration, the (sampling) weight $w_r$ for the $r$th feature is updated to be $\frac{w_r+\alpha_r\epsilon_m}{\sum_{v=1}^{p}(w_v+\alpha_v\epsilon_m)}$, where $\alpha_r$ is a pre-determined importance value of the $r$th feature. All the other steps are the same as those in \cite{FROSSYNIOTIS2004641}.

\subsection{Ensemble Selection}
Given the initial clustering ensemble $\mathscr{C}$, ensemble selection is mainly concerned with selecting a subset of base clusterings to form a final ensemble that achieves better performance. \cite{Fern:2008:CES} propose to partition the base clusterings into multiple groups by applying spectral clustering to the pair-wise NMI (sqrt) matrix. Each group is a subset of clusterings that are considered to be similar to one another. Then, for each group, select the clustering having the highest quality. The quality of clustering $C_m$ w.r.t. its group is defined as $\sum_{C\in\mathscr{C}}\phi(C_m, C)\delta(\mbox{$C_m$ and $C$ are in the same group}).$ The CSPA is applied to the selected clusterings to produce the final consensus clustering. \cite{Azimi:2009:ACE:} utilize the initial consensus clustering $C^*$ (obtained by applying the average linkage agglomerative clustering on the similarity matrix $\frac{1}{M}\mathbf{B}\mathbf{B}'$) to quantify the diversity of each base clustering. The diversity of clustering $C_m$ is $1-\phi(C_m, C^*)$. The ensemble $\mathscr{C}$ is said to be non-stable if the average diversity $1-\frac{1}{M}\sum_{m=1}^{M}\phi(C_m, C^*)$ is larger than 0.5. If the ensemble is non-stale, clusterings with diversity values higher than the median are selected to form the final ensemble.
\cite{FACELI20102809} adopt the corrected Rand index to rank the base clusterings. With reference to Table \ref{CT}, the corrected Rand index between $C$ and $C_m$ is
\begin{equation*}
\mbox{cR}(C, C_m)=\frac{\sum_{v=1}^{k}\sum_{r=1}^{k_m}{n_{vr}\choose2}-{n\choose2}^{-1}\sum_{v=1}^{k}{n^v\choose2}\sum_{r=1}^{k_m}{n_m^r\choose2}} {\frac{1}{2}[\sum_{v=1}^{k}{n^v\choose2}+\sum_{r=1}^{k_m}{n_m^r\choose2}]-{n\choose2}^{-1}\sum_{v=1}^{k}{n^v\choose2}\sum_{r=1}^{k_m}{n_m^r\choose2}}.
\end{equation*}
Sort the clusterings in ascending order according to the averaged values $\frac{1}{M}\sum_{C\in\mathscr{C}}\mbox{cR}(C, C_m)$, $m=1, \ldots, M$. Repeat the following two steps until the initial ensemble is empty: (1) Select the first clustering in the initial ensemble as the reference and place the reference in the final ensemble. (2) Remove from the initial ensemble the clusterings whose corrected Rand indices w.r.t. the reference are larger than a threshold.
\cite{JIA20111456} generate the initial ensemble by multiple runs of spectral clustering under different parameter settings. They apply different clustering validation criteria to produce different rankings of the base clusterings. More specifically, for an individual validation criterion (say, the NMI), half of the base clusterings are randomly selected and combined by the CSPA to produce a benchmark clustering; the NMI values between the base clusterings and the benchmark clustering are used to produce one ranking of the clusterings. The different ranks of a clustering are averaged to give the final rank, and a pre-specified number of top-ranked clusterings are selected. The CSPA is again applied to the selected clusterings to produce the final consensus clustering.
Given multiple relative clustering validation indices, \cite{Naldi2013} propose different strategies for selecting base clusterings. A base clustering will be selected if it has (1) the highest value of one relative clustering validation index, (2) the highest value of the average of the relative clustering validation indices, (3) the highest value of the weighted average of the relative clustering validation indices, or (4) the highest diversity w.r.t. the selected clusterings. Each strategy generates a subset of clusterings and then a candidate consensus clustering. The candidate consensus clusterings and the initial consensus clustering $C^*$ are validated by the relative clustering validation indices. The clustering with the highest (combined or individual) validation value is the final consensus clustering.
\cite{Zheng:2014:FHE} develop two methods for hierarchical-clustering (dendrogram) selection, in which tree distances are employed to measure the similarities between different dendrograms. The first method uses a modified $k$-medoids algorithm (with the tree distances) to cluster the dendrograms and then selects the medoid of each cluster. The second method starts with the medoid of all the input dendrograms and selects a dendrogram that is as far from the medoid as possible. Then the method repeatedly selects a dendrogram to maximize the distance to the nearest of the dendrograms selected so far.
\cite{YU20143362} treat the base clusterings as new features of the data and apply four feature selection methods to generate four subsets of clusterings from the initial ensemble. The quality of a clustering is measured by the average of the distances between data points and their nearest cluster centres. Let $s_t$ denote the $t$th subset ($t=1,2,3,4$), and the quality values of the clusterings in $s_t$ are represented by a quality vector $\pmb{q}^t=(q^t_C: C\in s_t)$. The weight for the $t$th subset is $w^t=\frac{\frac{1}{|s_t|}\sum_{C\in s_t}w_C^t-\max\{\frac{1}{|s_t|}\sum_{C\in s_t}w_C^t: ~t=1,2,3,4\}}{\frac{1}{4}\sum_{t=1}^{4}\frac{1}{|s_t|}\sum_{C\in s_t}w_C^t}$, where $w_C^t=\exp(-\frac{q^t_C-\min\{\pmb{q}^t\}}{\langle\pmb{q}^t, \pmb{1}\rangle/|s_t|})$, and the weight for clustering $C_m$ is $w_m=\sum_{t=1}^4\delta(C_m\in s_t)w^tw_{C_m}^t$. The final ensemble contains a pre-determined number of clusterings having the highest weights. The Ncut algorithm is applied to the (unweighted) similarity matrix, constructed from the final ensemble, to obtain the consensus clustering.
In \cite{6897160}, all base clusterings have $k$ clusters. After aligning the labels of the base clusterings, let $\pmb{F}^m$ denote the membership matrix for clustering $C_m$:
\begin{equation*}
\pmb{F}^m_{ir}=\frac{\frac{1}{|\mathbb{C}_m^r|}\sum_{\pmb{x}_j\in\mathbb{C}_m^r}\|\pmb{x}_i-\pmb{x}_j\|_2} {\sum_{v=1}^{k}\frac{1}{|\mathbb{C}_m^v|}\sum_{\pmb{x}_j\in\mathbb{C}_m^v}\|\pmb{x}_i-\pmb{x}_j\|_2},~~~~i=1, \ldots, n, ~r=1, \ldots, k.
\end{equation*}
The diversity of clustering $C_m$ is defined as $1-\frac{1}{M-1}\sum_{C\in\mathscr{C}, C\neq C_m}\mbox{aR}(C_m, C)$, where $\mbox{aR}(C_m, C)$ is the adjusted Rand index between $C_m$ and $C$; the quality of clustering $C_m$ is defined as $\frac{\sum_{1\leq v,r\leq k}\|\pmb{c}_m^r-\pmb{c}_m^v\|_2}{\frac{k-1}{2}\sum_{r=1}^{k}\sum_{\pmb{x}_i\in\mathbb{C}_m^r}\|\pmb{x}_i-\pmb{c}_m^r\|_2}$.
The base clusterings are selected in turn according to their diversity values (with the most diverse clustering selected first). The membership matrices of the selected clusterings are summed up, and a temporary consensus clustering is constructed from the aggregated membership matrix. The selection process stops when the quality of the temporary consensus clustering starts to decrease, and the current temporary consensus clustering is the final consensus clustering.
Given one similarity measure (e.g., the NMI), \cite{kanawati2015ensemble} constructs a relative neighborhood graph in which the nodes are the base clusterings, and the similarity between two nodes is the similarity between the two clusterings. Then a community detection algorithm is applied to the graph, and the clustering having the highest quality (e.g., the average NMI) within each community is selected into the final ensemble. Given multiple similarity measures, \cite{7403717} define a multilayer network in which each layer is a relative neighborhood graph constructed under one similarity measure. A seed-centric community detection algorithm is applied to the multilayer network. Under different internal clustering validation criteria, the clusterings within each community are ranked via an ensemble-ranking approach, and the top-ranked clustering is selected into the final ensemble. The CSPA is applied to the final ensemble to produce the consensus clustering.
\cite{7323847} study the problem of semi-supervised clustering for high dimensional data. The initial ensemble is obtained by applying the E$^2$CP method \citep{LuECP2013} to randomly generated data sets via subspace sampling. They define two indices: global index and local index. The global index of a clustering is the sum of the within-cluster variance and the number of unsatisfied constraints, while the local index of a clustering w.r.t. a set of clusterings measures the similarity between them. The clustering in the initial ensemble having the lowest global index is first selected into the final ensemble. In general, the remaining clusterings in the initial ensemble are sorted in ascending order according to their local indices w.r.t. the selected clusterings in the final ensemble; then select the clustering in the initial ensemble that, if placed in the final ensemble, decreases the global index of the consensus clustering of the final ensemble. The consensus clustering is obtained by applying the Ncut algorithm to the (unweighted) similarity matrix.
\cite{PIVIDORI2016120} generate the ensemble $\mathscr{C}$ by running the $k$-means algorithm on the data $\mathbb{X}$ with different values of $k$, and, for each $k$ value, with different initializations. Then a hierarchical agglomerative clustering method is applied to the resulted NMI (sqrt) matrix to partition the base clusterings into different groups, denoted by $\{G_1, \ldots, G_t\}$. Each group $G_g$ ($1\leq g\leq t$) is then represented by a clustering that shares the most information with the group members: $\arg\max_{C}\sum_{C_m\in G_g}\phi(C, C_m)$. Define a similarity matrix $\pmb{S}$ for the groups, where the similarity between two groups is the NMI value between their representative clusterings. Finally, group weights are defined as
\begin{equation*}
w_g=\frac{|G_g|}{1+\exp(\alpha[\frac{1}{t-1}\sum_{l=1, l\neq g}^{t}\pmb{S}_{gl}-\frac{1}{t(t-1)}\sum_{g=1}^{t}\sum_{l=1, l\neq g}^{t}\pmb{S}_{gl}])}, ~~1\leq g\leq t.
\end{equation*}
For a given $\alpha$ value and the size of the final ensemble $\tilde{M}$, we randomly sample $\frac{w_g}{\sum_{g=1}^{t}w_g}\tilde{M}$ (round to the nearest integer) clusterings from group $G_g$ to form the final ensemble. Hence, when $\alpha>0$, the final ensemble is more diverse than $\mathscr{C}$; by contrast, when $\alpha<0$, the final ensemble has lower diversity than $\mathscr{C}$. An arbitrary consensus function is applied to the finial ensemble to obtain the consensus clustering.
\cite{7482850} combine bagging and the $k$-means algorithm to create the initial ensemble. Utilizing the structure information of clustering $C_m$ ($m=1, \ldots, M$), a Gaussian mixture model (GMM), denoted by $\Phi_m$, is fitted to the $m$th bootstrap sample. Then a distribution-based distance measure is employed to quantify the similarity between two GMMs. The similarity score for the $m$th GMM is $\exp(-\frac{1}{M-1}\sum_{t=1, t\neq m}^{M}\mbox{distance between $\Phi_m$ and $\Phi_t$})$. A pre-specified number of GMMs with, say, highest similarity scores, are selected. A hypergraph is constructed from the selected GMMs, in which the vertices are the mixture components (Gaussian distributions) of the GMMs, a hyperedge exists between a vertex and its nearest neighbors, and the hyperedge weight is the difference between the total degree of the hyperedges linking the vertex and the degree of the vertex. The normalized hypergraph cut algorithm is used to partition the hypergraph and obtain a GMM (i.e., the final consensus clustering). \cite{ZHAO2017150} adopt five internal validity indices and the NMI (sqrt) to respectively measure the quality (the sum of the five internal validity indices) and diversity of the base clusterings. The clustering having the highest quality value is firstly placed into the final ensemble. Then repeat the following two steps until the size of the final ensemble is satisfied: (1) The diversity of each base clustering w.r.t. the clusterings in the final ensemble is calculated and multiplied by its quality value. (2) The base clustering with the highest product is removed from the initial ensemble and placed into the final ensemble.
\cite{Ma202015129} apply the k-means algorithm on the data $\mathbb{X}$ to generate half of the base clusterings and the spectral clustering algorithm to generate the other half. The diversity between two clusterings $C$ and $C_m$ is $1-\phi(C, C_m)$, and the quality of a clustering $C$ is $\frac{1}{M}\sum_{m=1}^{M}\phi(C, C_m)$.  They then apply a clustering algorithm on the diversity matrix to partition the base clusterings into groups and select the clustering with the highest quality from each group. By applying different clustering algorithms on the diversity matrix (and each time selecting the highest-quality group member), they obtain multiple subsets of base clusterings; the final ensemble is the union of all the subsets. The Ncut algorithm is applied to the similarity matrix of the final ensemble to obtain the consensus clustering.
\cite{10.1111/coin.12267} define an $M\times M$ matrix $\pmb{Q}$, where the diagonal entries are the quality measurements of the base clusterings, and the off-diagonal entries are the diversity measurements among base clusterings. They propose to solve the optimization problem $\max_{\pmb{\beta}\in\{0, 1\}^M}\pmb{\beta}'\pmb{Q}\pmb{\beta}-\lambda\|\pmb{\beta}\|_1$, which is then relaxed to optimizing a difference-of-convex function. The base clustering $C_m$ is selected if $\beta_m$ is larger than a threshold.

Instead of selecting base clusterings, a few research studies focus on selecting clusters. \cite{Alizadeh:2014:CES:} define a criterion for measuring the stability of a cluster. The stability of a cluster $\mathbb{C}\in C\in\mathscr{C}$ w.r.t. clustering $C_m$ is defined as
\begin{equation*}
\mbox{stability}(\mathbb{C}, C_m)=\frac{2|\mathbb{C}|\log(\frac{|\mathbb{C}|}{n})}{|\mathbb{C}|\log(\frac{|\mathbb{C}|}{n})+\sum_{r=1}^{k_m}|\mathbb{C}\cap\mathbb{C}_m^r|\log(\frac{|\mathbb{C}\cap\mathbb{C}_m^r|}{n})},
\end{equation*}
and the stability of the cluster w.r.t. the ensemble is defined as $\mbox{stability}(\mathbb{C})=\frac{1}{M}\sum_{m=1}^{M}\mbox{stability}(\mathbb{C}, C_m)$. The final consensus clustering is obtained by applying the average linkage agglomerative clustering on the similarity matrix $\pmb{S}$:
\begin{equation*}
\pmb{S}_{ij}=\frac{\sum_{m=1}^{M}\sum_{r=1}^{k_m}\delta(\pmb{x}_i\in\mathbb{C}_m^r)\delta(\pmb{x}_j\in\mathbb{C}_m^r)\delta(\mbox{stability}(\mathbb{C}_m^r)>\alpha)} {\max\{\sum_{m=1}^{M}\sum_{r=1}^{k_m}\delta(\pmb{x}_i\in\mathbb{C}_m^r)\delta(\mbox{stability}(\mathbb{C}_m^r)>\alpha), \sum_{m=1}^{M}\sum_{r=1}^{k_m}\delta(\pmb{x}_j\in\mathbb{C}_m^r)\delta(\mbox{stability}(\mathbb{C}_m^r)>\alpha)\}},
\end{equation*}
where $\alpha$ is a pre-specified threshold. \cite{Tang:2018:CEM:} define two indices for a cluster: a quantity index and a splitting index. Given a cluster $\mathbb{C}$, the authors construct an undirected graph in which the weight of an edge is the Euclidean distance between the two vertices; the minimum spanning tree for the graph is then generated. Let $\{w_1, \ldots, w_{|\mathbb{C}|-1}\}$ denote the edge weights of the minimum spanning tree, sorted in ascending order. The quantity index for the cluster is defined as $\frac{|\mathbb{C}|}{w_{|\mathbb{C}|-1}}$, and the splitting index is defined as
\begin{equation*}
\max\{\frac{w_{t+1}-w_t}{w_t}: t=1, \ldots, |\mathbb{C}|-2\}\times\frac{\arg\max\limits_{w_{t+1}}\{\frac{w_{t+1}-w_t}{w_t}: t=1, \ldots, |\mathbb{C}|-2\}-\frac{1}{|\mathbb{C}|-1}\sum_{t=1}^{|\mathbb{C}|-1}w_t} {\frac{1}{|\mathbb{C}|-1}\sum_{t=1}^{|\mathbb{C}|-1}w_t}.
\end{equation*}
Repeat the following four steps until the selected clusters cover all the objects in $\mathbb{X}$: (1) Identify the cluster with the highest quantity index, say $\mathbb{C}$. (2) If the splitting index of $\mathbb{C}$ is smaller than a threshold, $\mathbb{C}$ is selected and removed from the ensemble. (3) If the splitting index of $\mathbb{C}$ is larger than the threshold, $\mathbb{C}$ is divided into two sub-clusters by cutting the edge $\arg\max_{w_{t+1}}\{\frac{w_{t+1}-w_t}{w_t}: t=1, \ldots, |\mathbb{C}|-2\}$ in the corresponding minimum spanning tree; calculate the quantity and splitting indices for the sub-clusters and perform the splitting recursively until the splitting index is smaller than the threshold or the splitting depth reaches a threshold. (4) Include the sub-clusters into the ensemble. The similarity between objects $\pmb{x}_i$ and $\pmb{x}_j$ is the maximum of the quantity indices for the clusters that contain both $\pmb{x}_i$ and $\pmb{x}_j$. The consensus clustering is obtained by applying spectral clustering to the resulted similarity matrix.
The cluster reliability idea by \cite{Nazari2017} in Section \ref{FWC} is adopted by \cite{Abbasi20191311} to measure the ``stability'' of clusters, where the stability of a cluster $\mathbb{C}$ w.r.t. the ensemble $\mathscr{C}$ is defined as
\begin{equation*}
\mbox{stability}(\mathbb{C}, \mathscr{C})=\frac{1}{M}\sum_{m=1}^{m}\phi(\{\mathbb{C}, \mathbb{X}\setminus\mathbb{C}\}, \{P_{C_m}(\mathbb{C}), \mathbb{X}\setminus P_{C_m}(\mathbb{C})\}),
\end{equation*}
where $P_{C_m}(\mathbb{C})=\cup_{r=1}^{k_m}\mathbb{C}_m^r\delta(|\mathbb{C}_m^r\cap\mathbb{C}|>\frac{1}{2}|\mathbb{C}_m^r|)$. Only clusters whose stabilities are larger than a threshold are selected, and the following consensus clustering is done by extracting a co-association matrix from the selected clusters along with a linkage clustering.

\section{Variable-Weight Approach}\label{variableweights}
\subsection{Weighting Clusterings}
Based on the interpretation of the co-occurrence matrix $\mathbf{B}_m\mathbf{B}_m'$, an intuitive idea for clustering ensemble is to minimize the overall difference between the consensus clustering $C^*$ and the base clusterings:
\begin{equation}
\min\limits_{\mathbf{B}^*}~\frac{1}{M}\sum_{m=1}^{M}\|\mathbf{B}^*\mathbf{B}^{*'}-\mathbf{B}_m\mathbf{B}_m'\|^2_F, ~~~~\mbox{ s.t. } \mathbf{B}^*\in\{0, 1\}^{n\times k}, ~\mathbf{B}^*\pmb{1}=\pmb{1}.
\label{NMF0}
\end{equation}
Problem (\ref{NMF0}) is equivalent to
\begin{equation*}
\min\limits_{\mathbf{B}^*}~\|\mathbf{B}^*\mathbf{B}^{*'}-\frac{1}{M}\sum_{m=1}^{M}\mathbf{B}_m\mathbf{B}_m'\|^2_F, ~~~~\mbox{ s.t. } \mathbf{B}^*\in\{0, 1\}^{n\times k}, ~\mathbf{B}^*\pmb{1}=\pmb{1}.
\end{equation*}
The constraint that $\mathbf{B}^*$ is a binary membership indicator matrix can be relaxed to simplify the optimization. Specifically, by defining $\tilde{\mathbf{B}}^*=\mathbf{B}^*(\mathbf{B}^{*'}\mathbf{B}^*)^{-1/2}$, problem (\ref{NMF0}) can be relaxed into
\begin{equation}
\min\limits_{\mathbf{D}, \tilde{\mathbf{B}}^*} ~\|\tilde{\mathbf{B}}^*\mathbf{D}\tilde{\mathbf{B}}^{*'}-\frac{1}{M}\sum_{m=1}^{M}\mathbf{B}_m\mathbf{B}_m'\|^2_F, ~~~~\mbox{ s.t. } \tilde{\mathbf{B}}^{*'}\tilde{\mathbf{B}}^*=\mbox{diag}(\mathbf{1}), ~\tilde{\mathbf{B}}^*\geq0, ~\mathbf{D}\geq0, ~\mathbf{D}\mbox{ diagonal}.
\label{relax}
\end{equation}
\cite{LI2008} replace the simple average $\frac{1}{M}\sum_{m=1}^{M}\mathbf{B}_m\mathbf{B}_m'$ with a weighted average:
\begin{align}
\min_{\{w_m\}_{m=1}^M, \mathbf{B}^*}~&~\|\mathbf{B}^*\mathbf{B}^{*'}-\sum_{m=1}^{M}w_m\mathbf{B}_m\mathbf{B}_m'\|^2_F,\label{illpose}\\
\mbox{s.t.}~&~\mathbf{B}^*\in\{0, 1\}^{n\times k}, ~\mathbf{B}^*\pmb{1}=\pmb{1}, ~\sum_{m=1}^{M}w_m=1, w_m\geq0, ~m=1, \ldots, M\nonumber.
\end{align}
An iterative procedure is developed, in which the optimization over $\mathbf{B}^*$ while fixing $\{w_m\}_{m=1}^M$ is relaxed to a similar problem as (\ref{relax}). The optimization over $\{w_m\}_{m=1}^M$ while fixing $\mathbf{B}^*$ is a convex optimization problem. However, we remark that problem (\ref{illpose}) is ill-posed: $\forall$ $1\leq m\leq M$, the trivial solution with $w_m=1$ and $\mathbf{B}^*=\mathbf{B}_m$ is optimal. \cite{Lourenco2013} introduce two types of regularization on the clustering weights $\{w_1, \ldots, w_M\}$ to avoid putting the maximal weight on a single clustering:
\begin{align*}
\min_{\{w_m\}_{m=1}^M, \mathbf{B}^*}~&~\sum_{m=1}^{M}w_m\|\mathbf{B}^*\mathbf{B}^{*'}-\mathbf{B}_m\mathbf{B}_m'\|^2_F,\\
\mbox{s.t.}~&~\mathbf{B}^*\in\{0, 1\}^{n\times k}, ~\mathbf{B}^*\pmb{1}=\pmb{1}, ~\sum_{m=1}^{M}w_m=1, ~0\leq w_m\leq\lambda, ~m=1\ldots, M,
\end{align*}
and
\begin{align*}
\min_{\{w_m\}_{m=1}^M, \mathbf{B}^*}~&~\sum_{m=1}^{M}w_m\|\mathbf{B}^*\mathbf{B}^{*'}-\mathbf{B}_m\mathbf{B}_m'\|^2_F+\lambda\sum_{m=1}^{M}w_m^2,\\
\mbox{s.t.}~&~\mathbf{B}^*\in\{0, 1\}^{n\times k}, ~\mathbf{B}^*\pmb{1}=\pmb{1}, ~\sum_{m=1}^{M}w_m=1, ~w_m\geq0, ~m=1\ldots, M.
\end{align*}
Then the integer constraint $\mathbf{B}^*\in\{0, 1\}^{n\times k}$ is relaxed to be $\mathbf{B}^*\geq0$. An alternating, local optimization procedure is developed which interleaves updates of $\mathbf{B}^*$ and updates of $\{w_1, \ldots, w_M\}$.

Utilizing non-negative matrix factorization (NMF), \cite{Du:2011:CEV} approximate the concatenated matrix $\mathbf{B}$ by two low-rank matrices $\mathbf{U}$ and $\mathbf{V}$. The approximation is obtained by solving
\begin{align*}
\min_{\mathbf{U}, \mathbf{V}, \{w_m\}_{m=1}^M}~&~\|\mathbf{B}-\mathbf{U}\mathbf{V}'\|^2_F +\frac{\lambda}{2}\sum_{m=1}^{M}w_m\sum_{i,j=1}^{n} [\mathbf{B}_m\mathbf{B}_m']_{ij}\|\mathbf{U}_{i\cdot}-\mathbf{U}_{j\cdot}\|^2_2,\\
\mbox{s.t.}~&~\mathbf{U}\geq0, ~\mathbf{V}\geq0, ~\sum_{m=1}^{M}w_m=1, ~w_m\geq0, ~m=1, \ldots, M,
\end{align*}
or
\begin{align*}
\min_{\mathbf{U}, \mathbf{V}, \{w_m\}_{m=1}^M}~&~\sum_{i, v}(\mathbf{B}_{iv}\log(\frac{\mathbf{B}_{iv}}{[\mathbf{U}\mathbf{V}']_{iv}})-\mathbf{B}_{iv}+[\mathbf{U}\mathbf{V}']_{iv}) +\frac{\lambda}{2}\sum_{m=1}^{M}w_m\sum_{i,j=1}^{n} [\mathbf{B}_m\mathbf{B}_m']_{ij}[\mbox{KL}(\mathbf{U}_{i\cdot}\|\mathbf{U}_{j \cdot})+\mbox{KL}(\mathbf{U}_{j\cdot}\|\mathbf{U}_{i\cdot})],\\
\mbox{s.t.}~&~\mathbf{U}\geq0, ~\mathbf{V}\geq0, ~\sum_{m=1}^{M}w_m=1, ~w_m\geq0, ~m=1, \ldots, M,
\end{align*}
where $\mbox{KL}(\cdot, \cdot)$ represents the Kullback-Leibler divergence. $\lambda$ controls the smoothness of the low-rank matrix $\mathbf{U}$. The above two problems are solved by iteratively updating $\{w_1, \ldots, w_M\}$, $\mathbf{U}$ and $\mathbf{V}$. The optimal low-rank matrix $\mathbf{U}$ is then used to extract the consensus clustering.

\cite{Ou-Yang2013} combine weighted clustering ensemble, Bayesian probability and NMF to identify protein complexes. Let $\mathbf{F}\in\mathbb{R}^{n\times k}$ denote the true unknown protein-complex propensity matrix, where $\mathbf{F}_{iv}$ is the probability of object $\pmb{x}_i$ (a protein) belonging to the $v$th cluster (a complex). Hence, $[\mathbf{F}\mathbf{F}']_{ij}$ measures the probability of proteins $\pmb{x}_i$ and $\pmb{x}_j$ belonging to the same complex. They assume that, given $[\mathbf{F}\mathbf{F}']_{ij}$, $[\mathbf{BWB}']_{ij}$ has a Poisson distribution with rate $[\mathbf{F}\mathbf{F}']_{ij}$. Hence, the likelihood of the weighted similarity matrix is $\Pr(\mathbf{BWB}'|\mathbf{F})=\prod_{i,j=1}^n \frac{([\mathbf{F}\mathbf{F}']_{ij})^{[\mathbf{BWB}']_{ij}}}{\Gamma([\mathbf{BWB}']_{ij}+1)} \exp(-[\mathbf{F}\mathbf{F}']_{ij})$. They then assume that $\{\mathbf{F}_{1v}, \mathbf{F}_{2v}, \ldots, \mathbf{F}_{nv}\}$ are i.i.d. and follow a truncated normal distribution with parameter $\beta_v$. Finally, the $k$ parameters $\{\beta_1, \beta_2, \ldots, \beta_k\}$ are assumed to be i.i.d. and follow an inverse Gamma distribution. Write $\pmb{\beta}=(\beta_1, \beta_2, \ldots, \beta_k)$. The objective is to maximize the penalized log-likelihood:
\begin{align*}
&\max_{\{w_m\}_{m=1}^M, \mathbf{F}, \pmb{\beta}}\log(\Pr(\mathbf{BWB}', \mathbf{F}, \pmb{\beta}))+\lambda H(w_1, \ldots, w_M)\\
&=\max_{\{w_m\}_{m=1}^M, \mathbf{F}, \pmb{\beta}} \log(\Pr(\mathbf{BWB}'|\mathbf{F}))+\log(\Pr(\mathbf{F}|\pmb{\beta}))+\log(\Pr(\pmb{\beta}))+\lambda H(w_1, \ldots, w_M),
\end{align*}
subject to the constraints that $\mathbf{F}\geq0$, $\mathbf{F}\pmb{1}=\pmb{1}$, $\pmb{\beta}\geq0$, $\sum_{m=1}^{M}w_m=1$, $w_m\geq0$, $m=1, \ldots, M$. An iterative algorithm is developed, in which the clustering weight $w_m$ has an explicit expression. The optimal solution of $\mathbf{F}$ is used to identify the consensus clustering (protein complexes).

For text clustering, \cite{Hongfu2015} first employ a random feature sampling scheme to deal with the high-dimensional problem, and then apply the $k$-means algorithm on the generated subsets to create the clustering ensemble. With reference to the contingency table (Table \ref{CT}), the consensus clustering is the optimal solution of the following problem:
\begin{align*}
\min_{\{w_m\}_{m=1}^M, C}~&~-\sum_{m=1}^{M}w_m[H(C_m)-\sum_{r=1}^{k}\frac{n^r}{n}H(\frac{n_{r1}}{n^r}, \ldots, \frac{n_{rk_m}}{n^r})] +\lambda\sum_{m=1}^{M}w_m^2,\\
\mbox{s.t.}~&~C\in\mathscr{C}_{\mathbb{X}}, \sum_{m=1}^{M}w_m=1, w_m\geq0, m=1, \ldots, M.
\end{align*}
The quantity within the brackets is therein called utility, with a large value indicating strong similarity between $C$ and $C_m$. With the weights fixed, the optimization over $C\in\mathscr{C}_{\mathbb{X}}$ can be equivalently transformed into an information-theoretic $k$-means clustering, where the distance matric is the Kullback-Leibler divergence. With $C$ fixed, the authors employ a coordinate descent-based method to optimize the weights.

The kernel k-means algorithm transfers the original data into a high-dimensional or infinite-dimensional space in which the cluster are separated from each other. The objective function of the kernel k-means algorithm is
\begin{align*}
\min_{\mathbf{B}\in\{0, 1\}^{n\times k}} ~~\mbox{trace}(\mathbf{K})-\mbox{trace}(\mbox{diag}(\mathbf{1}'\mathbf{B})^{-\frac{1}{2}}\mathbf{B}'\mathbf{K}\mathbf{B}\mbox{diag} (\mathbf{1}'\mathbf{B})^{-\frac{1}{2}}),~~\mbox{s.t.}~~\mathbf{B}\mathbf{1}=\mathbf{1},
\end{align*}
where $\mathbf{K}\in\mathbb{R}^{n\times n}$ is the kernel matrix. By defining $\mathbf{F}=\mathbf{B}\mbox{diag}(\mathbf{1}'\mathbf{B})^{-\frac{1}{2}}$, the problem can be relaxed into $\min_{\mathbf{F}\in\mathbb{R}^{n\times k}} ~~\mbox{trace}(\mathbf{K}(\mathbf{I}-\mathbf{F}\mathbf{F}')),~~\mbox{s.t.}~~\mathbf{F}'\mathbf{F}=\mathbf{I}.$
\cite{8832148} apply $M$ different kernel functions to obtain the $M$ base clusterings, with clustering $C_m$ represented by the clustering matrix $\mathbf{F}^m$. The objective is to maximally align the consensus clustering matrix with the rotated base clustering matrices:
\begin{align*}
\min_{\mathbf{F}^*, \{w_m, \mathbf{R}_m\}_{m=1}^M} ~&~\mbox{trace}(\mathbf{F}^{*'}\sum_{m=1}^{M}w_m\mathbf{F}^m\mathbf{R}_m)+\lambda_1\mbox{trace}(\mathbf{F}^{*'}\mathbf{F}^0)- \lambda_2\sum_{m=1}^{M}\sum_{v=1}^{M}w_mw_v\mbox{trace}(\mathbf{R}_{m'}\mathbf{F}^{m'}\mathbf{F}^v\mathbf{R}_v),\\
\mbox{s.t.}~&~\mathbf{F}^{*'}\mathbf{F}^*=\mathbf{I}, \mathbf{R}_m'\mathbf{R}_m=\mathbf{I}, \sum_{m=1}^{M}w_m=1, ~w_m\geq0, ~m=1\ldots, M,
\end{align*}
where $\mathbf{R}_m\in\mathbb{R}^{k\times k}$ is a rotation matrix, and $\mathbf{F}^0$ is a given prior clustering matrix. The trace value $\mbox{trace}(\mathbf{R}_{m'}\mathbf{F}^{m'}\mathbf{F}^v\mathbf{R}_v)$ measures the correlation between the rotated matrices $\mathbf{F}^m\mathbf{R}_m$ and $\mathbf{F}^v\mathbf{R}_v$.

\cite{LI2020105337} replace the traditional Frobenius norm with the Bregman divergence and relax the co-occurrence matrix $\mathbf{B}_m\mathbf{B}_m'$ to a similarity matrix $\mathbf{S}^m$, where $0\leq\mathbf{S}^m_{ij}\leq1$. The optimization problem is
\begin{align*}
\min_{\{w_m\}_{m=1}^M, \mathbf{S}^*}~&~\sum_{m=1}^{M}w_m\sum_{i=1}^{n}\sum_{j=1}^{n}D_{\varphi}(\mathbf{S}^*_{ij}, \mathbf{S}^m_{ij})
+\lambda\sum_{m=1}^{M}w_m^2,\\
\mbox{s.t.}~&~\mathbf{S}^*=\mathbf{S}^{*'}, \mathbf{S}^*\geq0, ~\sum_{m=1}^{M}w_m=1, ~w_m\geq0, ~m=1\ldots, M,
\end{align*}
where $\varphi$ is a strictly convex function, and $D_{\varphi}(\cdot, \cdot)$ is the Bregman divergence: $D_{\varphi}(x_1, x_2)=\varphi(x_1)-\varphi(x_2)-\nabla\varphi(x_2)(x_1-x_2)$. A block coordinate descent algorithm is developed to minimize the above objective. The method is used to analyze the similarity between molecules in mRNA expression data, DNA methylation data, or microRNA expression data, to further study subtypes of cancer.

\subsection{Weighting Clusters}
\cite{ZheX:2015} introduce weights over clusters in the form of $\mathbf{B}\odot\pmb{W}$, where $\pmb{W}=[\pmb{W}_1, \ldots, \pmb{W}_M]\in\mathbb{R}^{n\times\sum_{m=1}^{M}k_m}$ and $(\mathbf{B}\odot\pmb{W})\mathbf{1}=\mathbf{1}$. Here and in the following, the operator $\odot$ represents the element-wise multiplication. The objective function is
\begin{align*}
\min_{\mathbf{B}^*, \pmb{W}, \mathbf{P}}~&~\|\mathbf{B}\odot\pmb{W}-\mathbf{B}^*\mathbf{P}\|^2_F+\lambda\|\pmb{W}\|^2_F,\\
\mbox{s.t.}~&~\mathbf{B}^*\in\{0, 1\}^{n\times k}, ~\mathbf{B}^*\pmb{1}=\pmb{1}, \pmb{W}\geq0, (\mathbf{B}\odot\pmb{W})\mathbf{1}=\mathbf{1}, \mathbf{P}\in\{0, 1\}^{|C^*|\times\sum_{m=1}^{M}k_m},
\end{align*}
where $\mathbf{P}=[\mathbf{P}_1, \ldots, \mathbf{P}_M]$, and $\mathbf{P}_m\in\{0, 1\}^{|C^*|\times k_m}$ is a permutation matrix to align the columns of $\mathbf{B}^*$ so that $\mathbf{B}_m\odot\pmb{W}_m$ should be well approximated by $\mathbf{B}^*\mathbf{P}_m$. The problem is then transformed into an NMF problem by relaxing $\mathbf{B}^*$ and $\mathbf{P}$, and an iterative algorithm is developed to update $\{\mathbf{B}^*, \mathbf{P}\}$ and $\pmb{W}$ alternately.

\subsection{Weighting Features}\label{VWF}
For a base clustering $C=\{\mathbb{C}^1, \ldots, \mathbb{C}^k\}\in\mathscr{C}$, define a set of weight vectors $\{\pmb{w}_1, \ldots, \pmb{w}_k\}$, where $\pmb{w}_v=(w_{v1}, \ldots, w_{vp})$ is the weight vector for cluster $\mathbb{C}^v$ ($v=1, \ldots, k$), $w_{vr} (\geq0)$ is the weight for feature $r$ ($r=1, \ldots, p$), and $\sum_{r=1}^p w_{vr}=1$.
In \cite{Muna2006} and \cite{Domeniconi:2009}, given the value of $\lambda$, $C$ is obtained by minimizing the following error:
\begin{equation}
\min_{\{\pmb{c}^v, \pmb{w}_v\}_{v=1}^k}~\sum_{v=1}^{k}\frac{1}{|\mathbb{C}^v|}\sum_{\pmb{x}_i\in \mathbb{C}^v}\langle\pmb{w}_v, (\pmb{c}^v-\pmb{x}_i)^2\rangle-\lambda\sum_{v=1}^{k} H(\pmb{w}_v), ~~\mbox{ s.t. } \pmb{w}_v\geq0, \|\pmb{w}_v\|_1=1, ~v=1, \ldots, k.
\label{entropy}
\end{equation}
The regularization term $\lambda H(\pmb{w}_v)$ penalizes solutions putting the maximal weight on the single feature that has the smallest dispersion within cluster $\mathbb{C}^v$. They develop an algorithm (called the locally adaptive clustering algorithm) that iteratively optimizes over $\{\pmb{c}^v\}_{v=1}^k$ and $\{\pmb{w}_v\}_{v=1}^k$ until convergence. After clustering $C$ is obtained, for each object $\pmb{x}_i$, the authors define a probability mass function over the clusters $\{\mathbb{C}^1, \ldots, \mathbb{C}^k\}$: $(\Pr(\pmb{x}_i\in\mathbb{C}^1), \ldots, \Pr(\pmb{x}_i\in\mathbb{C}^k))$, which is calculated from the weighted distances $\langle\pmb{w}_v, (\pmb{c}^v-\pmb{x}_i)^2\rangle$ for $v=1, \ldots, k$. A new similarity matrix is defined in which the similarity between objects $\pmb{x}_i$ and $\pmb{x}_j$ is the cosine similarity between their probability mass functions $(\Pr(\pmb{x}_i\in\mathbb{C}^1), \ldots, \Pr(\pmb{x}_i\in\mathbb{C}^k))$ and $(\Pr(\pmb{x}_j\in\mathbb{C}^1), \ldots, \Pr(\pmb{x}_j\in\mathbb{C}^k))$. An ensemble of $M$ clusterings (and hence $M$ similarity matrices $\{\mathbf{S}_1, \ldots, \mathbf{S}_M\}$) are obtained by taking $M$ different values of $\lambda$. Finally, the METIS is applied on the resulting graph $G(\frac{1}{M}\sum_{m=1}^{M}\mathbf{S}_m)$ to compute the consensus clustering that minimizes the edge weight-cut.

After obtaining the clustering ensemble via the locally adaptive clustering algorithm, \cite{6890436} modify the similarity matrices $\{\mathbf{S}_1, \ldots, \mathbf{S}_M\}$ by considering the similarity between two clusters. Given the set of probability mass functions $\{(\Pr(\pmb{x}_i\in\mathbb{C}^1_m), \ldots, \Pr(\pmb{x}_i\in\mathbb{C}^{k_m}_m)): m=1, \ldots, M, i=1, \ldots, n\}$, the similarity between clusters $\mathbb{C}^r_{m_1}$ and $\mathbb{C}^v_{m_2}$ from different clusterings is defined as $s(\mathbb{C}^r_{m_1}, \mathbb{C}^v_{m_2})=\frac{\sum_{\pmb{x}_i\in\mathbb{X}}\min\{\Pr(\pmb{x}_i\in\mathbb{C}^r_{m_1}), ~\Pr(\pmb{x}_i\in\mathbb{C}^v_{m_2})\}}{\sum_{\pmb{x}_i\in\mathbb{X}}\max\{\Pr(\pmb{x}_i\in\mathbb{C}^r_{m_1}), ~\Pr(\pmb{x}_i\in\mathbb{C}^v_{m_2})\}}$,
from which we can calculate the similarity $s(\mathbb{C}^r_m, \mathbb{C}^v_m)$ between clusters $\mathbb{C}^r_m$ and $\mathbb{C}^v_m$ from the same clustering, which is the normalized value of $\sum_{\tilde{\mathbb{C}}\in\mathscr{N}(\mathbb{C}^r_m)\cap\mathscr{N}(\mathbb{C}^v_m)} \frac{s(\tilde{\mathbb{C}}, \mathbb{C}^r_m)+s(\tilde{\mathbb{C}}, \mathbb{C}^v_m)}{\sum_{\mathbb{C}\in\mathscr{N}(\tilde{\mathbb{C}})} s(\tilde{\mathbb{C}}, \mathbb{C})}$. Here, $\mathscr{N}(\mathbb{C}^r_m)$ is the set of clusters that are neighbors of cluster $\mathbb{C}^r_m$: if $\tilde{\mathbb{C}}\cap\mathbb{C}^r_m\neq\varnothing$, then $\tilde{\mathbb{C}}\in\mathscr{N}(\mathbb{C}^r_m)$. Denote $\mathbb{C}_m^{i*}=\arg\max_{\mathbb{C}^r_m\in C_m}\Pr(\pmb{x}_i\in\mathbb{C}^r_m)$. Then the probability mass function $(\Pr(\pmb{x}_i\in\mathbb{C}_m^1), \ldots, \Pr(\pmb{x}_i\in\mathbb{C}_m^{k_m}))$ is modified to be
\begin{equation*}
\Pr(\pmb{x}_i\in\mathbb{C}^r_m)\leftarrow\Pr(\pmb{x}_i\in\mathbb{C}_m^{i*})\times s(\mathbb{C}_m^{i*}, \mathbb{C}^r_m),~~~~r=1, \ldots, k.
\end{equation*}
The similarity matrix $\mathbf{S}_m$ is calculated from the modified probability mass functions $\{(\Pr(\pmb{x}_i\in\mathbb{C}^1_m), \ldots, \Pr(\pmb{x}_i\in\mathbb{C}^{k_m}_m)): i=1, \ldots, n\}$. Finally, spectral clustering is applied on $\frac{1}{M}\sum_{m=1}^{M}\mathbf{S}_m$ to determine the consensus clustering.

\cite{Parvin2013} and \cite{Parvin2015} generalize the locally adaptive clustering algorithm by introducing another weight vector $\pmb{\omega}=(\omega_1, \ldots, \omega_k)$ for weighting the clusters $\{\mathbb{C}^1, \ldots, \mathbb{C}^k\}$. Given the values of the parameters $\lambda_1$ and $\lambda_2$, the base clustering $C$ is obtained by solving
\begin{align*}
\min_{\pmb{\omega}, \{\pmb{c}^v, \pmb{w}_v\}_{v=1}^k}~&~\sum_{v=1}^{k}\frac{\omega_v}{|\mathbb{C}^v|}\sum_{\pmb{x}_i\in \mathbb{C}^v}\langle\pmb{w}_v, (\pmb{c}^v-\pmb{x}_i)^2\rangle-\lambda_1\sum_{v=1}^{k}H(\pmb{w}_v)-\lambda_2H(\pmb{\omega}),\\
\mbox{s.t.}~&~\pmb{\omega}\geq0, ~\|\pmb{\omega}\|_1=1, ~\pmb{w}_v\geq0, ~\|\pmb{w}_v\|_1=1, ~v=1, \ldots, k,
\end{align*}
for hard clustering, and
\begin{align*}
\min_{\pmb{\omega}, \{\pmb{c}^v, \pmb{w}_v\}_{v=1}^k, \mathbf{F}}~&~\sum_{v=1}^{k}\omega_v\left[\sum_{i=1}^n(\mathbf{F}_{iv})^{\alpha}\langle\pmb{w}_v, (\pmb{c}^v-\pmb{x}_i)^2\rangle-\lambda_1H(\pmb{w}_v)\right]-\lambda_2H(\pmb{\omega}),\\
\mbox{s.t.}~&~\mathbf{F}\geq0, ~\mathbf{F}\pmb{1}=\pmb{1}, ~\pmb{\omega}\geq0, ~\|\pmb{\omega}\|_1=1, ~\pmb{w}_v\geq0, ~\|\pmb{w}_v\|_1=1, ~v=1, \ldots, k,
\end{align*}
for fuzzy clustering. $\mathbf{F}\in\mathbb{R}^{n\times k}$, and $\mathbf{F}_{iv}$ is the fuzzy membership of object $\pmb{x}_i$ to cluster $\mathbb{C}^v$. Likewise, they develop an algorithm that iteratively optimizes over $\{\pmb{c}^v\}_{v=1}^k$, $\{\pmb{w}_v\}_{v=1}^k$, $\pmb{\omega}$, and $\mathbf{F}$ until convergence. An ensemble of $M$ clusterings is obtained by varying the values of $\lambda_1$ and $\lambda_2$, and finally a consensus clustering is extracted out of the ensemble by traditional methods, e.g., the CSPA.

In \cite{CORDEIRODEAMORIM201762}, the ensemble $\mathscr{C}=\{C_1, \ldots, C_M\}$ is created by the Minkowski weighted $k$-means with the Minkowski exponent $\rho$ taking $M$ different values. For a clustering $C=\{\mathbb{C}^1, \ldots, \mathbb{C}^k\}$ with the set of weight vectors $\{\pmb{w}_1, \ldots, \pmb{w}_k\}$, the objective function of the Minkowski weighted $k$-means is
\begin{equation*}
\min_{\{\pmb{c}^v, \pmb{w}_v\}_{v=1}^k}\sum_{v=1}^{k}\sum_{\pmb{x}_i\in\mathbb{C}^v} \|\pmb{w}_v\odot(\pmb{x}_i-\pmb{c}^v)\|^\rho_\rho, ~~~~\mbox{ s.t. } \pmb{w}_v\geq0, ~\|\pmb{w}_v\|_1=1, ~v=1, \ldots, k.
\end{equation*}
Given the centroids $\{\pmb{c}^v\}_{v=1}^k$, the optimal weight $w_{vr}$ ($v=1, \ldots, k$ and $r=1, \ldots, p$) is inversely proportional to the dispersion of feature $r$ in cluster $\mathbb{C}^v$. For each value of the exponent $\rho$, the authors run the Minkowski weighted $k$-means for 100 times with random initializations. A base clustering is selected from the 100 runs with the maximum value of a clustering validation index, e.g., the Calinski-Harabasz index. An ensemble of 41 base clusterings is created by setting $\rho=1.0, 1.1, \ldots, 5.0$. The consensus clustering is selected from the 41 base clusterings having the highest sum of adjusted Rand indices between itself and the rest.

The algorithm developed by \cite{Zhou2017} simultaneously optimizes two objective functions, one accounting for intra-cluster compactness and the other accounting for inter-cluster separation. For a base clustering $C=\{\mathbb{C}^1, \ldots, \mathbb{C}^k\}$, the first objective function is
\begin{equation*}
\min_{\{\pmb{c}^v, \pmb{w}_v\}_{v=1}^k}\sum_{v=1}^{k}\sum_{\pmb{x}_i\in \mathbb{C}^v}\sum_{r=1}^{p}w_{vr}\times\mbox{kd}(x_{ir}, c^v_r), ~~\mbox{ s.t. } \pmb{w}_v\geq0, ~\|\pmb{w}_v\|_1=1, ~v=1, \ldots, k,
\end{equation*}
where $\mbox{kd}(x_{ir}, c^v_r)$ calculates the kernel distance between $\pmb{x}_i$ and $\pmb{c}^v$ on the $r$th dimension. The second objective function is
\begin{equation*}
\min_{\{\pmb{c}^v, \pmb{w}_v\}_{v=1}^k}\frac{\sum_{v=1}^{k}\sum_{\pmb{x}_i\in \mathbb{C}^v}\sum_{r=1}^{p}w_{vr}\times\mbox{kd}(x_{ir}, c^v_r)}{\min_{t\neq v}\{\mbox{kd}(\pmb{c}^v, \pmb{c}^t)\}\times\sum_{v=1}^{k}\sum_{r=1}^{p}w_{vr}\times\mbox{kd}(\bar{x}_r, c^v_r)}, ~~\mbox{ s.t. } \pmb{w}_v\geq0, ~\|\pmb{w}_v\|_1=1, ~v=1, \ldots, k,
\end{equation*}
where $\bar{\pmb{x}}=\frac{1}{n}\sum_{i=1}^n\pmb{x}_i$. The feature weights and the centroids are updated iteratively. The clustering ensemble is created via subset sampling from $\mathbb{X}$, and the consensus clustering is obtained by a graph partitioning method.

\section{Conclusions and Future Work}\label{Conclusion}
Earlier works on clustering ensemble assume equal contribution of every clustering/cluster/feature/object to the ensemble. Theoretical analysis \citep[e.g., ][]{1410288} shows that, under certain conditions, the probability for the consensus clustering to uncover the intrinsic data structure increases as the ensemble size increases. In real-life clustering problems, however, ensemble size is always limited, and theoretical assumptions can be violated. It would be desirable to obtain the highest possible clustering quality with a limited ensemble size. With that end in view, a number of weighted clustering ensemble methods are suggested, aiming to speed up the convergence to the optimal clustering. In this paper, we explore main weighted clustering ensemble methods, taking into account the weighting mechanism as well as the mathematical and computational tools used by each method. The bibliographical compilation is presented in a unified framework, through a homogeneous exposition of weight recipients (clusterings, clusters, features, or objects) and weight properties (constant or variable).

Compared with the variable-weight approach, the computation of weight values in the fixed-weight approach is trivial. However, fixed weights should summarize as much information as possible from the given clustering ensemble. To this end, one could aggregate various clustering validation criteria into the weight values; alternatively, one could employ different types of weights. For example, both clustering weights and cluster weights can be incorporated into a weighted clustering ensemble method; that way, both clustering quality and cluster stability are considered. The variable-weight approach has the drawback of high computational complexity: the weight values often have to be determined iteratively, and the determination of $\lambda$ (i.e., the amount of regularization/penalization) is non-trivial. However, variable weights have the attractive interpretation that they are optimal w.r.t. a given objective function. We here remark that regularization/penalization in the variable-weight approach has a significant influence on the performance of the final clustering. For example, the local manifold regularization in \cite{7837980} will enforce the locally geometrical structure of the data in the final clustering. A related topic to clustering ensemble is community detection in (multilayer) networks \citep{Tagarelli2017}, to which the weighting paradigm summarized herein can be extended.

\begin{appendices}
\section{Applications to Distributed Data}\label{application}
\subsection{Multi-View Data Clustering}
In multi-view data clustering, objects are described by multiple views, each view providing its own light on the data. Multi-view data are often of multiple modalities or come from multiple sources. For example, user groups can be formed based on user profiles, user online social connections, user historical transactions, etc. Suppose we have a data set with $n$ objects and $M$ views $\{\mathbb{X}^m, m=1, \ldots, M\}$, where $\mathbb{X}^m\in\mathbb{R}^{n\times d_m}$ is the data matrix for view $m$. In the ensemble $\mathscr{C}=\{C_1, \ldots, C_M\}$, each base clustering is constructed from one view of the data.

\cite{ASI:ASI21312} propose two approaches to clustering journal papers, where the multiple views are from multiple data sources such as text mining data and citation data. In the first approach, the consensus clustering is obtained by applying the average linkage agglomerative clustering on the weighted similarity matrix $\mathbf{BWB}'$, where the clustering weight $w_m$ takes the normalized value of $\frac{1}{M-1}\sum_{C\neq C_m}\phi(C, C_m)$. In the second approach, the $M$ data matrices are first mapped into the same feature space by one feature mapping. Then the $M$ data matrices in the feature space are combined by weighted average, where the weights are still the clustering weights. Finally, a standard clustering algorithm, e.g., the kernel $k$-means algorithm, is applied on the aggregated matrix to obtain the consensus clustering.

\cite{Shao2015} apply the clustering ensemble technique to incomplete multi-view data. The multi-view data are incomplete in that, for any view $m$, the data matrix $\mathbb{X}^m$ has a number of rows missing. They propose to fill the missing rows with estimated values yet assign lower weights: A binary indicator matrix $\mathbf{I}\in\mathbb{R}^{M\times n}$ is defined: $\mathbf{I}_{mi}=\delta(\mbox{the $m$th view of the $i$th object is available})$. For each view $m$, define a diagonal matrix of weights $\mathbf{W}^m\in\mathbb{R}^{n\times n}$ as $\mathbf{W}^m_{ii}=\delta(\mathbf{I}_{mi}=1)+\frac{\sum_{i=1}^{n}\mathbf{I}_{mi}}{n}\delta(\mathbf{I}_{mi}=0)$. Hence, in view $m$, the objects whose values are missing receive the same weight $\frac{1}{n}\sum_{i=1}^{n}\mathbf{I}_{mi}$. Using the idea of NMF, the problem is formulated as
\begin{align}
\min\limits_{\mathbf{U}^*, \{\mathbf{U}^m, \mathbf{V}^m\}_{m=1}^M}~&~\sum_{m=1}^{M}\left[\|\mathbf{W}^m(\mathbb{X}^m-\mathbf{U}^m \mathbf{V}^{m'})\|^2_F +\lambda_{1m}\|\mathbf{W}^m(\mathbf{U}^m-\mathbf{U}^*)\|^2_F +\lambda_{2m}\|\mathbf{U}^m\|_{2,1}\right],
\label{decomposable}\\
\mbox{s.t.}~&~\mathbf{U}^*\geq0, \mathbf{U}^m\geq0, \mathbf{V}^m\geq0, ~~m=1, \ldots, M,\nonumber
\end{align}
where $\|\cdot\|_{2,1}$ is the $L_{2,1}$ norm. $\mathbf{U}^*$ is called the consensus latent feature matrix. An alternating scheme is developed to optimize the objective function. The $k$-means algorithm is then applied on $\mathbf{U}^*$ to get the consensus clustering. Function (\ref{decomposable}) can be decomposed w.r.t. the objects, and hence \cite{7840701} further develop an online incomplete multi-view data clustering algorithm.

To reduce the influence from the noise and outliers in the data, \cite{7900020} replace the Frobenius norm in the NMF by the $L_{2,1}$ norm:
\begin{align*}
\min\limits_{\mathbf{U}, \{\mathbf{V}^m, w_m\}_{m=1}^M}~&~\sum_{m=1}^{M}\|\mathbb{X}^m-\mathbf{U}\mathbf{V}^{m'}\|_{2,1}+\lambda\sum_{m=1}^{M}w_m^{\alpha_1}\sum_{i\neq j}\pmb{A}^m_{ij}\|\mathbf{U}_{i\cdot}-\mathbf{U}_{j\cdot}\|_2^2,\\
\mbox{s.t.}~&~\mathbf{U}\geq0, \mathbf{U}'\mathbf{U}=\mbox{diag}(\mathbf{1}), \sum_{m=1}^Mw_m=1, w_m\geq0, \mathbf{V}^m\geq0,~~m=1, \ldots, M,
\end{align*}
where $\pmb{A}^m$ is the adjacency matrix for the $m$th view:
\begin{equation*}
\pmb{A}^m_{ij}=\exp(-\frac{1}{\alpha_2}\|\mathbb{X}^m_{i\cdot}-\mathbb{X}^m_{j\cdot}\|_2)\delta(\mathbb{X}^m_{i\cdot}\mbox{ is a $k$-nearest neighbor of }\mathbb{X}^m_{j\cdot},\mbox{ or vice versa}).
\end{equation*}
The second term in the above objective function is called the local manifold regularization. NMF only considers the globally geometrical structure of the data space, whereas the local manifold regularization can preserve the locally geometrical structure of the data space. The largest element in $\mathbf{U}_{i\cdot}$ indicates the cluster of object $\pmb{x}_i$ in the consensus clustering. \cite{TANG201849} incorporate feature selection into the NMF framework:
\begin{align*}
\min_{\mathbf{U}^*, \{\mathbf{U}^m, \mathbf{V}^m, \gamma_m, \beta_m, w_m\}_{m=1}^M}~&~\sum_{m=1}^{M}\left[\|\mathbb{X}^m-\mathbf{U}^m\mathbf{V}^{m'}\|_F^2+\gamma_m^{\alpha_1}\|\mathbf{V}^m\|_{2,1}+\beta_m^{\alpha_2}\|\mathbf{U}^m-\mathbf{U}^*\|^2_F +w_m^{\alpha_3}\sum_{i\neq j}\pmb{A}^m_{ij}\|\mathbf{U}^m_{i\cdot}-\mathbf{U}^m_{j\cdot}\|_2^2\right],\\
\mbox{s.t.}~&~\mathbf{U}^*\geq0, \mathbf{U}^{*'}\mathbf{U}^*=\mbox{diag}(\mathbf{1}), \sum_{m=1}^{M}\gamma_m=\sum_{m=1}^{M}\beta_m=\sum_{m=1}^{M}w_m=1,\\
~&~\mathbf{U}^m\geq0, \mathbf{V}^m\geq0, \gamma_m\geq0, \beta_m\geq0, w_m\geq0,~~m=1, \ldots, M.
\end{align*}
The $k$-means algorithm is applied on the selected features to obtain the consensus clustering.

Other than NMF, \cite{7837980} apply concept factorization for multi-view data clustering. Concept factorization aims to approximate $\mathbb{X}^m$ by $\mathbf{U}^m \mathbf{V}^{m'}\mathbb{X}^m$, where $\mathbf{V}^m$ is called the association matrix indicating the degree of $\mathbb{X}^m$ related to the concepts, and $\mathbf{U}^m$ is called the projection/representation matrix denoting the projection of $\mathbb{X}^m$ onto the concepts. Concept factorization has the property that the matrices $\mathbf{U}^m$ and $\mathbf{U}^m$ both tend to be very sparse. The regularized concept factorization problem is formulated as
\begin{align*}
\min_{\mathbf{U}^*, \{\mathbf{U}^m, \mathbf{V}^m, w_m\}_{m=1}^M}~&~\sum_{m=1}^{M}\|\mathbb{X}^m-\mathbf{U}^m \mathbf{V}^{m'}\mathbb{X}^m\|_F^2+\lambda_1\sum_{m=1}^{M} w_m\|\mathbf{U}^m-\mathbf{U}^*\|^2_F+\lambda_2\sum_{m=1}^{M}\sum_{i\neq j}\pmb{A}^m_{ij}\|\mathbf{U}^m_{i\cdot}-\mathbf{U}^m_{j\cdot}\|^2_2,\\
\mbox{s.t.}~&~\mathbf{U}^*\geq0, \sum_{m=1}^{M}w_m=1, w_m\geq0, \mathbf{U}^m\geq0, \mathbf{V}^m\geq0,~~m=1, \ldots, M,
\end{align*}
where the adjacency matrix $\pmb{A}^m$ is defined as $\pmb{A}^m_{ij}=\exp(-\frac{\|\mathbb{X}^m_{i\cdot}-\mathbb{X}^m_{j\cdot}\|^2_2}{\max\{\mbox{var}(\mathbb{X}^m_{i\cdot}), \mbox{var}(\mathbb{X}^m_{j\cdot})\}})$. The manifold regularization is again to encode the local geometrical structure of the data. The optimization variables are updated iteratively. The largest element of $\mathbf{U}^*_{i\cdot}$ indicates the cluster of object $\pmb{x}_i$ in the consensus clustering.

The multi-view data studied in \cite{DECARVALHO2015115} are represented by a set of $M$ relational matrices: $\{\mathbf{R}_1, \ldots, \mathbf{R}_M\}$, where $[\mathbf{R}_m]_{ij}$ measures the dissimilarity between objects $\pmb{x}_i$ and $\pmb{x}_j$ from the $m$th view. Assume that all the base clusterings have $k$ clusters, and that each cluster has a representative element from $\mathbb{X}$, called the medoid. Let $\pmb{c}_m^r$ denote the medoid of the cluster $\mathbb{C}_m^r$, $r=1, \ldots, k$ and $m=1, \ldots, M$. The fuzzy consensus clustering is characterized by the fuzzy membership matrix $\mathbf{F}\in\mathbb{R}^{n\times k}$ which is obtained by solving
\begin{align}
\min_{\mathbf{F}, \{\pmb{c}_m^r, w_{mr}: r=1, \ldots, k\}_{m=1}^M}~&~\sum_{m=1}^{M}\sum_{r=1}^{k}w_{mr}\sum_{i=1}^{n}(\mathbf{F}_{ir})^\alpha \sum_{j=1}^{n} [\mathbf{R}_m]_{ij}\delta(\pmb{c}_m^r=\pmb{x}_j),\label{relation}\\
\mbox{s.t.}~&~\mathbf{F}\geq0, ~\mathbf{F}\pmb{1}=\pmb{1}, \prod_{m=1}^{M}w_{mr}=1, w_{mr}>0, \pmb{c}_m^r\in\mathbb{X}, ~~r=1, \ldots, k, m=1, \ldots, M.\nonumber
\end{align}
Initially, $\pmb{c}_m^r$ is randomly selected from $\mathbb{X}$, and $w_{mr}=1$ for $r=1, \ldots, k$ and $m=1, \ldots, M$; given $\{\pmb{c}_m^r: r=1, \ldots, k\}_{m=1}^M$ and $\{w_{mr}: r=1, \ldots, k\}_{m=1}^M$, $\mathbf{F}$ is determined by minimizing (\ref{relation}). Then the algorithm iteratively updates $\{\pmb{c}_m^r: r=1, \ldots, k\}_{m=1}^M$, $\{w_{mr}: r=1, \ldots, k\}_{m=1}^M$ and $\mathbf{F}$.

For multi-view data fuzzy clustering, \cite{WANG2017457} propose to minimize the maximal sum of weighted disagreements of different views. Assuming that all base clusterings have $k$ clusters, the objective function is
\begin{align}
\min_{\mathbf{F}, \{\pmb{c}_m^r: r=1, \ldots, k\}_{m=1}^M}~&~\max\limits_{\{w_m\}_{m=1}^M}~\sum_{m=1}^{M}w_m^{\alpha_1}\sum_{r=1}^{k}\sum_{i=1}^{n} (\mathbf{F}_{ir})^{\alpha_2}\|\mathbb{X}^m_{i\cdot}-\pmb{c}_m^r\|_2^2,\label{feature}\\
\mbox{s.t.}~&~\mathbf{F}\geq0, ~\mathbf{F}\pmb{1}=\pmb{1}, \sum_{m=1}^{M}w_m=1, w_m\geq0, ~~m=1, \ldots, M,\nonumber
\end{align}
where $\mathbf{F}\in\mathbb{R}^{n\times k}$, and $\mathbf{F}_{ir}$ is the fuzzy membership of the $i$th object to the $r$th cluster. The parameter $\alpha_1$ controls the distribution of the weights, while $\alpha_2$ controls the fuzziness of the membership. The objective function is convex w.r.t. $\mathbf{F}$ and $\{\pmb{c}_m^r: r=1, \ldots, k\}_{m=1}^M$, and is concave w.r.t. $\{w_m\}_{m=1}^M$. Therefore, an alternating optimization procedure is developed. The consensus clustering is then generated from the optimal fuzzy membership matrix $\mathbf{F}$.

As with \cite{DECARVALHO2015115}, \cite{ZHAO2020105459} also study the cluster-weight optimization problem. Assume that all base clusterings have $k$ clusters. The objective involves three aspects. The first aspect is to learn the base clusterings using the idea of NMF: $\min_{\mathbf{U}^m\geq0, \mathbf{V}^m\geq0}~~\|\mathbb{X}^m-\mathbf{U}^m \mathbf{V}^{m'}\|^2_F$, and the cluster label of the $i$th object is determined by the index of the largest entry in the coefficient vector $\mathbf{U}^m_{i\cdot}$. The second aspect is to learn the cluster weights by solving the following problem: $\min_{\mathbf{U}^*\geq0, \mathbf{W}^m\geq0}~\sum_{m=1}^{M}\|\mathbf{U}^m-\mathbf{U}^*\mathbf{W}^m\|^2_F$, s.t. $\sum_{m=1}^{M}\mathbf{W}^m=\mbox{diag}(\mathbf{1})$. Here, $\mathbf{W}^m$ is the $k\times k$ diagonal matrix of cluster weights. The third aspect is to give more weights to clusters that are more distant from the other, by minimizing $\sum_{m=1}^{M}\sum_{v=1}^{k}(1-\mathbf{W}^m_{vv})\sum_{r=1}^{k}\|\mathbf{V}^m_{v\cdot}-\mathbf{V}^m_{r\cdot}\|_2$. By combining all the three aspects, the overall objective function is
\begin{align*}
\min_{\mathbf{U}^*, \{\mathbf{U}^m, \mathbf{V}^m, \mathbf{W}^m\}_{m=1}^M}~&~\sum_{m=1}^{M}\|\mathbb{X}^m-\mathbf{U}^m \mathbf{V}^{m'}\|^2_F+ \lambda_1\sum_{m=1}^{M}\|\mathbf{U}^m-\mathbf{U}^*\mathbf{W}^m\|^2_F+\\ &~~~~~~~~\lambda_2\sum_{m=1}^{M}\sum_{v=1}^{k}(1-\mathbf{W}^m_{vv})\sum_{r=1}^{k}\|\mathbf{V}^m_{v\cdot}-\mathbf{V}^m_{r\cdot}\|_2+ \lambda_3\mathbf{1}'\mathbf{W}^m\mathbf{W}^m\mathbf{1},\\
\mbox{s.t.}~&~\mathbf{U}^m\geq0, \mathbf{V}^m\geq0, \mathbf{U}^*\geq0, \mathbf{W}^m\geq0, \sum_{m=1}^{M}\mathbf{W}^m=\mbox{diag}(\mathbf{1}).
\end{align*}
An iterative procedure is developed to optimize the objective function.

\cite{KANG2020279} apply the subspace clustering technique to create the ensemble: the graph construction step involves minimizing $\|\mathbb{X}^m-\mathbf{Z}^{m}\mathbb{X}^m \|^2_F+ \lambda\|\mathbf{Z}^{m}\|^2_F$, where $\mathbf{Z}^{m}\in\mathbb{R}^{n\times n}$ is the affinity graph matrix; the spectral clustering step involves minimizing trace$(\mathbf{F}^{m'}(\mbox{diag}(\mathbf{Z}^{m}\mathbf{1})-\mathbf{Z}^{m})\mathbf{F}^m)$. To integrate clustering weights into consensus clustering, the overall objective function is
\begin{align*}
\min_{\mathbf{F}^{*}, \{\mathbf{Z}^m, \mathbf{F}^m\}_{m=1}^M}~&~\sum_{m=1}^{M}\{\|\mathbb{X}^m-\mathbf{Z}^{m}\mathbb{X}^m \|^2_F+ \lambda_1\|\mathbf{Z}^{m}\|^2_F+\lambda_2\mbox{trace}(\mathbf{F}^{m'}(\mbox{diag}(\mathbf{Z}^{m}\mathbf{1})-\mathbf{Z}^{m})\mathbf{F}^m)+\\
&~~~~~~~~\lambda_3w_m\|\mathbf{F}^*\mathbf{F}^{*'}-\mathbf{F}^{m}\mathbf{F}^{m'}\|^2_F\},\\
\mbox{s.t.}~&~\mathbf{F}^{m'}\mathbf{F}^{m}=\mbox{diag}(\mathbf{1}), \mathbf{F}^{*'}\mathbf{F}^{*}=\mbox{diag}(\mathbf{1}), \mathbf{Z}^m\geq0.
\end{align*}
Although the clustering weights are not optimization parameters, in the optimization procedure once $\mathbf{F}^*$ is updated, the weights are re-calculated according to $w_m=\frac{1}{2\|\mathbf{F}^*\mathbf{F}^{*'}-\mathbf{F}^{m}\mathbf{F}^{m'}\|_F}$.

\subsection{Temporal Data Clustering}
For temporal data clustering, a representation-based algorithm can convert temporal data clustering into static data clustering via a parsimonious representation. However, one single representation tends to encode only those features well presented in its own representation space and inevitably incurs information loss. \cite{Yang2011} employ $M$ different representation-based algorithms to create $M$ different static data sets from the original temporal data. Then an arbitrary clustering algorithm is employed to create $M$ different base clusterings, one base clustering from one static data set. When the ensemble $\mathscr{C}=\{C_1, \ldots, C_M\}$ is available, a consensus clustering is obtained by applying an average linkage hierarchical clustering method on the weighted similarity matrix $\mathbf{B}\mathbf{W}\mathbf{B}'$, where the weight $w_m$ for clustering $C_m$ is the value of a clustering validation index, e.g. the $\mbox{DVI}$. The authors apply different clustering validation criteria to obtain different consensus clusterings. Finally, these consensus clusterings are treated as base clusterings, and the average linkage hierarchical clustering method is applied on the resulted un-weighted similarity matrix to get the final consensus clustering.

In \cite{YANG2018391}, the $M$ base clusterings are obtained by applying the HMM-based k-model on the temporal data $\mathbb{X}$ under $M$ different initializations. Each cluster $\mathbb{C}^r_m$ in clustering $C_m$ is modelled by one hidden Markov model (HMM); clustering $C_m$ is represented by a mixture of $k_m$ HMMs, and the likelihood of object $\pmb{x}_i$ conditioned on clustering $C_m$ is
\begin{equation*}
\Pr(\pmb{x}_i|C_m)=\sum_{r=1}^{k_m}\Pr(\mathbb{C}^r_m)\Pr(\pmb{x}_i|\mathbb{C}^r_m) =\sum_{r=1}^{k_m}\Pr(\mbox{HMM for }\mathbb{C}^r_m)\Pr(\pmb{x}_i|\mbox{HMM for }\mathbb{C}^r_m).
\end{equation*}
$\Pr(\mbox{HMM for }\mathbb{C}^r_m)$ is a given prior probability. A bi-weighting scheme is proposed to weight both clusterings and clusters. Clustering weights take into account clustering quality, while cluster weights take into account cluster size.
Particularly, the weight $w^r_m$ for cluster $\mathbb{C}^r_m$ is $\frac{|\mathbb{C}^r_m|}{n}$, and the weight $w_m$ for clustering $C_m$ is the normalized value of
\begin{equation*}
\exp(\frac{\sum_{r=1}^{k_m}\Pr(\mbox{HMM for }\mathbb{C}^r_m)\times\mbox{KL}\left(\Pr(\mathbb{X}|\mbox{HMM for }\mathbb{C}^r_m)\|\Pr(\mathbb{X}|C_m)\right)}{\alpha}),
\end{equation*}
where $\Pr(\mathbb{X}|\mbox{HMM for }\mathbb{C}^r_m)$ and $\Pr(\mathbb{X}|C_m)$ are probability mass functions obtained by respectively normalizing $\{\Pr(\pmb{x}_i|\mbox{HMM for }\mathbb{C}^r_m)\}_{i=1}^n$ and $\{\Pr(\pmb{x}_i|C_m)\}_{i=1}^n$. The consensus clustering is obtained by applying a hierarchical agglomerative clustering method on the weighted similarity matrix $\mathbf{B}\mathbf{W}\mathbf{B}'$ in which $\mathbf{W}=\mbox{diag}(\underbrace{w_1w_1^1, \ldots, w_1w_1^{k_1}}\limits_{k_1}, \ldots, \underbrace{w_Mw_M^1, \ldots, w_Mw_M^{k_M}}\limits_{k_M})$.

\end{appendices}

\bibliographystyle{apalike}
\let\OLDthebibliography\thebibliography
\renewcommand\thebibliography[1]{
  \OLDthebibliography{#1}
  \setlength{\parskip}{0pt}
  \setlength{\baselineskip}{0.67cm}
  \setlength{\itemsep}{0cm}
}
\bibliography{Eref}

\end{document}